\documentclass{article}

\usepackage[preprint]{neurips_2021}

\usepackage[utf8]{inputenc} 
\usepackage[T1]{fontenc}    
\usepackage{hyperref}       
\usepackage{url}            
\usepackage{booktabs}       
\usepackage{amsfonts}       
\usepackage{nicefrac}       
\usepackage{microtype}      
\usepackage{xcolor}         
\usepackage{bbm}
\usepackage{amsfonts}       
\usepackage{nicefrac}       
\usepackage{microtype}  
\usepackage{amsmath}
\usepackage{varwidth}
\usepackage{lscape} 
\usepackage{graphicx}
\usepackage{adjustbox}
\usepackage{wrapfig}
\usepackage{overpic}

\usepackage{xcolor}
\usepackage{color}
\definecolor{deepblue}{rgb}{0,0,0.5}
\definecolor{deepred}{rgb}{0.6,0,0}
\definecolor{deepgreen}{rgb}{0,0.5,0}
\definecolor{darkgreen}{rgb}{0,0.6,0}
\definecolor{darkred}{rgb}{0.7,0.0,0}
\definecolor{darkblue}{rgb}{0,0.0,0.6}
\definecolor{magenta}{rgb}{0.8,0.1,0.8}
\definecolor{darksomething}{rgb}{0,0.4,0.6}

\newcommand{\dd}{{\mathrm{d}}}
\newcommand{\pd}[2]{\frac{\partial #1}{\partial #2}}

\newcommand{\od}[2]{\frac{\dd #1}{\dd #2}}

\title{Gradients are Not All You Need}

\author{%
  Luke Metz\thanks{Equal contribution} \quad C. Daniel Freeman\footnotemark[1] \quad Samuel S. Schoenholz \\
  Google Research, Brain Team\\
  \texttt{\{lmetz, cdfreeman, schsam\}@google.com} \\
  \AND
  Tal Kachman \\
  Radboud University \\ 
  Donders Institute for Brain, Cognition and Behaviour\\
  \texttt{tal.kachman@donders.ru.nl}
}

\begin{document}

\maketitle

\begin{abstract}
    Differentiable programming techniques are widely used in the community and are responsible for the machine learning renaissance of the past several decades.
    While these methods are powerful, they have limits.
    In this short report, we discuss a common chaos based failure mode which appears in a variety of differentiable circumstances, ranging from recurrent neural networks and numerical physics simulation to training learned optimizers.
    We trace this failure to the spectrum of the Jacobian of the system under study, and provide criteria for when a practitioner might expect this failure to spoil their  differentiation based optimization algorithms.
\end{abstract}

\section{Introduction}
Owing to the overwhelming success of deep learning techniques in providing fast function approximations to almost every problem practitioners care to look at, it has become popular to try to make differentiable implementations of different systems---the logic being, that by taking the tried-and-true suite of techniques leveraging derivatives when optimizing neural networks for a task, one need only take their task of interest, make it differentiable, and place it in the appropriate place in the pipeline and train ``end to end''.
This has lead to a plethora of differentiable software packages, ranging across rigid body physics \citep{heiden2021neuralsim,hu2019taichi,werling2021fast,degrave2019differentiable,de2018end, gradu2021deluca,freeman2021brax}, graphics \citep{li2018differentiable, kato2020differentiable}, molecular dynamics \citep{jaxmd2020, hinsen2000molecular}, differentiating though optimization procedures \citep{maclaurin2016modeling}, weather simulators \citep{bischof1996sensitivity}, and nuclear fusion simulators \citep{mcgreivy2021optimized}.

Automatic differentiation provides a conceptually straightforward handle for computing derivatives though these systems, and often can be applied with limited compute and memory overhead \citep{paszke2017automatic,ablin2020super,margossian2019review,bischof1991exploiting,corliss2013automatic}.
The resulting gradients however, while formally ``correct'' in the sense that they are exactly the desired mathematical object \footnote{Up to numerical precision, though see \citep{chow1992numerical,kachman2017numerical} for cases where the gradients will be close.}, might not be algorithmically useful---especially when used to optimize certain functions of system dynamics. In this work, we discuss one potential issue that arises when working with iterative differentiable systems: chaos.

Chaotic dynamics and difficulties differentiating through them are not a discovery of this work. To our knowledge, discussion first appeared in climate modeling \citep{lea2000sensitivity, kohl2002adjoint}, but have expanded to a varity of different domains ranging from neural network initialization and activation design \citep{yang2017mean, hayou2019impact}, model based control \citep{parmas2018pipps, Parmas_phdthesis}, meta-learning \citep{metz2019understanding}, fluid simulation \citep{ni2017sensitivity, kochkov2021machine}, and learning protein structures \citep{ingraham2018learning}.

\section{Preliminaries: Iterated Dynamical Systems}
Chaos emerges naturally in iterated maps \citep{bischof1991exploiting,ruelle2009review}.  Consider the following discrete matrix equation:
\begin{equation}
s_{k+1} = A_k s_k
\end{equation}
where $A_k$ is some possibly state-dependent matrix describing how the state information, $s_k$, transforms during a step.
It is not difficult to show (see App. \ref{Appendix}), under appropriate assumptions, that the trajectories of $x_k$ depend on the eigenspectrum of the family of transformations $A_k$.
Crucially, if the largest eigenvalue of the $A_k$ is typically larger than 1, then trajectories will tend to diverge exponentially like the largest eigenvalue $\lambda_{max}^k$. 
Contrariwise, if the largest eigenvalue is less than one, trajectories will tend to vanish.

Of course, dynamical systems encountered in the wild are usually more complicated, so suppose instead that we have a transition function, $f$ which depends on state data, $s$, and control variables $\theta$:  
\begin{equation}
s_{t+1} = f(s_t, \theta)
\end{equation}
We're typically concerned with functions of our control variables, evaluated over a trajectory, for example, consider some loss function which sums losses ($l_t$) computed up to a finite number of steps $N$:
\begin{equation}
l(\theta) =  \sum^N_{t=0} l_t(s_t, \theta).
\label{eq:all_ml}
\end{equation}
This formalism is extremely general, and equation \ref{eq:all_ml} encompasses essentially the entire modern practice of machine learning. For several common examples of $f, s, \theta$, and $l$, see Table \ref{tab:examples}.
\begin{table}
\caption{Different kinds of machine learning techniques often resemble differentiating through some iterative system. $f$ is the iteratively applied function, $s$ is an input, $\theta$ stands for parameters and $l$ is the optimization objective \label{tab:examples}}
\begin{adjustbox}{center}
\begin{tabular}{|l|l|l|l|l|}
\hline
\textbf{Domain}         & \textbf{$f$}                                                                          & \textbf{$s$}                                                                              & \textbf{$\theta$}                                                                                              & \textbf{$l$}                                                                                                                                               \\ \hline
Neural Network Training & \begin{tabular}[c]{@{}l@{}}A layer transformation \\ of a neural network\end{tabular} & \begin{tabular}[c]{@{}l@{}}The inputs to\\  that layer\end{tabular}                       & \begin{tabular}[c]{@{}l@{}}The weight matrix \\ and bias vector \\ for that layer\end{tabular}                 & \begin{tabular}[c]{@{}l@{}}cross entropy \\ mean squared error, \\ l2 regularization, \\ etc.\end{tabular}                                                                           \\ \hline
Reinforcement Learning  & \begin{tabular}[c]{@{}l@{}}The step function \\ of an environment\end{tabular}        & \begin{tabular}[c]{@{}l@{}}The state data of \\ the environment \\ and agent \end{tabular}              & \begin{tabular}[c]{@{}l@{}}The parameters\\  of a policy\end{tabular}                                          & \begin{tabular}[c]{@{}l@{}}The reward \\ function for \\ the environment\end{tabular}                                                                      \\ \hline
Learned Optimization    & \begin{tabular}[c]{@{}l@{}}The application of \\ an optimizer\end{tabular}            & \begin{tabular}[c]{@{}l@{}}The parameters \\ in a network \\ being optimized\end{tabular} & \begin{tabular}[c]{@{}l@{}}The tunable \\ parameters for \\ the optimizer, \\ e.g., learning rate\end{tabular} & \begin{tabular}[c]{@{}l@{}}The performance \\ of the network \\ being optimized \\ on a task after \\ some number of \\ steps of optimization\end{tabular} \\ \hline
\end{tabular}

\end{adjustbox}
\end{table}

Solving our problem usually amounts to either maximizing or minimizing our loss function, $l$, and a differentiably minded practitioner is usually concerned with the derivative of $l$.  Consider the derivatives of the first few terms of this sum:
\begin{align}
\od{l_0}{\theta} &= \pd{l_0}{s_0} \pd{s_0}{\theta} + \pd{l_0}{\theta}\\
\od{l_1}{\theta} &= \pd{l_1}{s_1}\pd{s_1}{s_0} \pd{s_0}{\theta} + \pd{l_1}{s_1} \pd{s_1}{\theta} + \pd{l_1}{\theta} \\
\od{l_2}{\theta} &= \pd{l_2}{s_2}\pd{s_2}{s_1}\pd{s_1}{s_0} \pd{s_0}{\theta} + \pd{l_2}{s_2}\pd{s_2}{s_1} \pd{s_1}{\theta} +  \pd{l_2}{s_2} \pd{s_2}{\theta}  + \pd{l_2}{\theta}
\end{align}
Here, the pattern is conceptually clear, and then for an arbitrary $t$:
\begin{equation}
\od{l_t}{\theta} = \pd{l_t}{\theta} + \sum_{k=1}^{t} \pd{l_t}{s_t} \left(\prod_{i=k}^{t} \pd{s_i}{s_{i-1}} \right) \pd{s_k}{\theta}
\end{equation}

This leaves us with a total loss:
\begin{equation}
    \dfrac{\dd l}{\dd \theta} = \frac{1}{N} \sum_{t=0}^N \left[ \pd{l_t}{\theta} + \sum_{k=1}^{t} \pd{l_t}{s_t} \left(\prod_{i=k}^{t} \pd{s_i}{s_{i-1}} \right) \pd{s_k}{\theta} \right]
    \label{eq:product}
\end{equation}

Note the product $\left(\prod_{i=k}^{t} \pd{s_i}{s_{i-1}} \right)$ appearing on the right hand side of equation \ref{eq:product}.  The matrix of partial derivatives $\pd{s_i}{s_{i-1}}$ is exactly the Jacobian of the dynamical system ($f$), and this has precisely the iterated structure discussed in the beginning of this section.
Thus, one might not be surprised to find that the gradients of loss functions of dynamical systems depend intimately on the spectra of Jacobians.

At the risk of belaboring the point: As $N$ grows, the number of products in the sum also grows.
If $\pd{s_i}{s_{i-1}}$ is a constant then the gradient will either exponentially grow, or shrink in $N$ leading to exploding or vanishing gradients.
When the magnitude of all eigenvalues of $\pd{s_i}{s_{i-1}}$ are less than one, the system is stable and the resulting product will be well behaved.
If some, or all, of the eigenvectors are above one, the dynamics can diverge, and could even be chaotic \citep{bollt2000controlling}.
If the underlying system is known to be chaotic, namely small changes in initial conditions result in diverging states -- e.g. rigid body physics simulation in the presence of contacts \citep{coluci2005chaotic} -- this product will diverge. This concept is often colloquially called the butterfly effect \citet{lorenz1963deterministic}.

\begin{wrapfigure}{r}{0.5\textwidth}
    \vspace{-2em}
    \centering
    \includegraphics[width=0.4\textwidth]{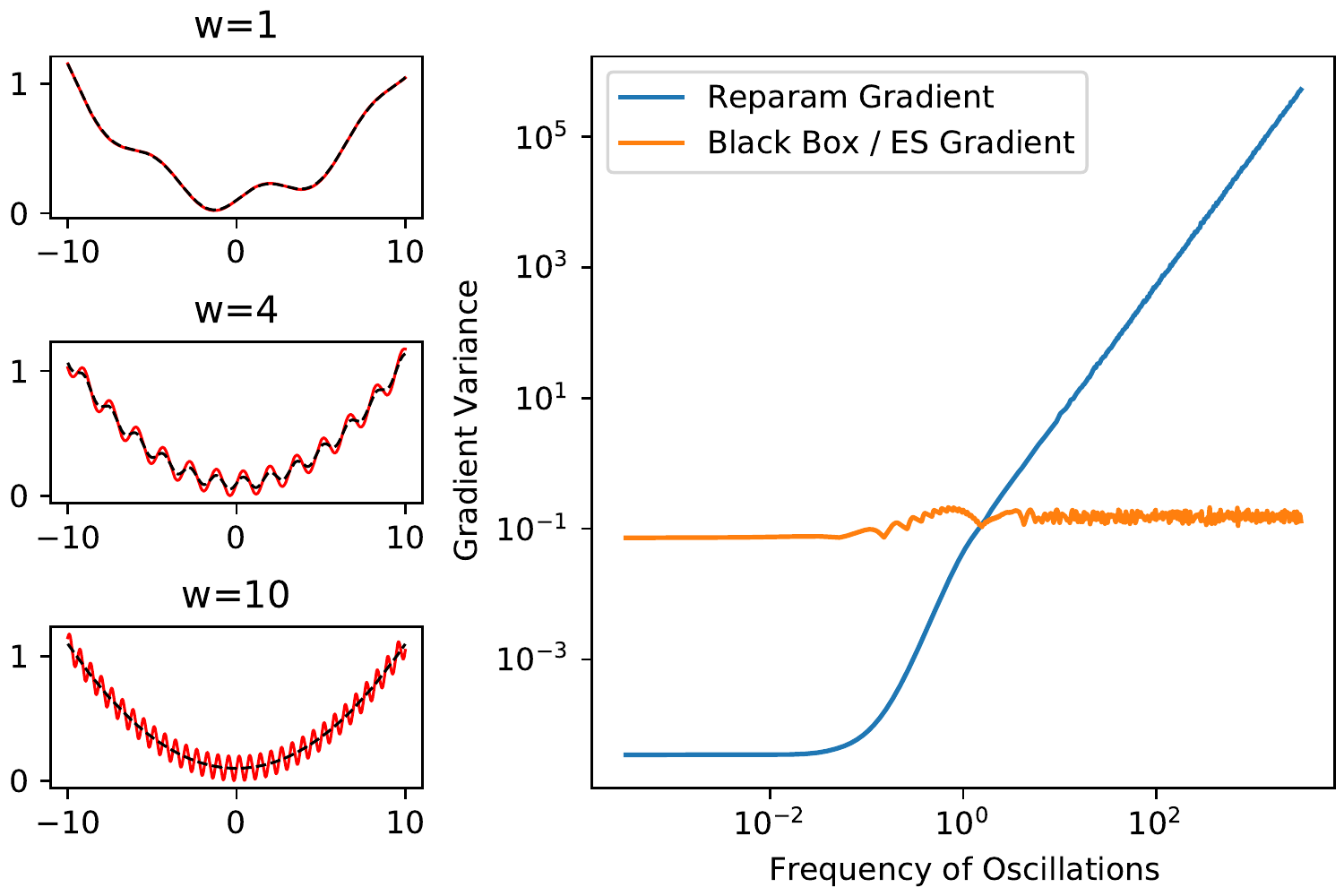}
    \vspace{-1em}
    \caption{Sometimes, black box gradient estimates can result in lower variance gradient estimates.
    On the left, we plot $l(x) = 0.1 \text{sin}(xw/(\pi)) + (x / 10)^2 + 0.1$ for different values of $w$ in red, as well as the loss smoothed by convolving with a 0.3 std Gaussian. On the figure to the right we show the max gradient variance computed over all $x \in [-10, 10]$. When the frequency of oscillations grows higher, the reparameterization gradient variance also grows while the back box gradient remains constant.
    \label{fig:reparm_es}}
\end{wrapfigure}

Thus far, we have made the assumption that the system is deterministic.
For many systems we care about the function $f$ is modulated by some, potentially stochastic, procedure.
In the case of neural network training and learned optimization, this randomness could come from different minibatches of data, in the case of reinforcement learning this might come from the environment, or randomness from a stochastic control policy.
In physics simulation, it can even come from floating point noise in how engine calculations are handled on an accelerator.
To compute gradients a combination of the reparameterization trick \citep{kingma2013auto} and Monte Carlo estimates are usually employed, averaging the gradients computed by backpropagating through the non-stochastic computations~\citep{schulman2015gradient}.
The resulting loss surface of such stochastic systems are thus ``smoothed'' by this stochasticity and, depending on the type of randomness, this could result in better behaved loss functions (e.g. smoother, meaning the true gradient has a smaller norm).
In chaotic systems, this notion of smoothing is related to the "chaotic hypothesis" which loosely states that time averages in ergodic system are well behaved even if individual trajectories are not \citep{ruelle1995measures, gallavotti1995dynamical}.
Due to the nature of how the reparameterized gradients are estimated -- taking products of sequences of state Jacobian matrices -- they can result in extremely high gradient norms.
In some cases, throwing out the fact that the underlying system is differentiable and using black box methods to estimate {\it the same} gradient can result in a better estimate of the exact same gradient \citep{parmas2018pipps, metz2019understanding}!  Even when the underlying objective is smooth, but chaotic dynamics and noise give rise to exploding gradients, these black box methods are known to provide low variance estimates, as pointed out by \citep{parmas2018pipps}.

As a simple example of this, consider some recurrent system with loss $\hat{l}(\theta)$ made stochastic by sampling parameter noise from a Gaussian between $\mathcal{N}(0, I \sigma^2)$ where $\sigma^2$ is the standard deviation of the smoothing. If the underlying loss is bounded in some way, then the smoothed loss's gradients will also be bounded, controlled by the amount of smoothing ($\sigma^2$). If using the reparameterization trick and backprop to estimate gradients of this this, however, depend on the unsmoothed loss $\hat{l}$, and could have extremely large gradients even growing exponentially and thus exponentially many samples will be needed to obtain accurate estimates. One can instead employ a black box estimate similar to evolutionary strategies \citep{rechenberg1973evolutionsstrategie, schwefel1977evolutionsstrategien, wierstra2008natural, schulman2015gradient} or variational optimization \citep{staines2012variational} to compute a gradient estimate. Because this just works with the function evaluations of the un-smoothed loss the gradient estimates again become better behaved. This comes at the cost, however, of poor performance with increased dimensionality.

The fact that black box gradients can have better variance properties is counter intuitive. As a pictorial demonstration, consider figure \ref{fig:reparm_es}. Instead of using a recurrent system, we simply plot $\hat{l}(x) = 0.1 \text{sin}(xw/(\pi)) + (x / 10)^2 + 0.1$ and vary $w$ as a proxy for potentially exploding gradients. We assume a Gaussian smoothing of this unsmoothed loss ($\sigma = 0.3$). We find for lower frequency oscillations the reparameterization gradient results in lower variance gradient estimates where as for higher frequency oscillations the black box estimate has lower variance gradients and the reparameterization variance continues to grow. A similar set of experiments comparing among gradient estimators has also been shown in \citet{gal2016uncertainty, mohamed2020monte}.
This example, and many others explored in this paper are multiscale in that they have a high curvature local structure, with some other global structure. \citet{kong2020stochasticity} showed that even even when performing noise free gradient descent, with a large enough learning rate, optimization resembles stochastic gradient descent converging to distributions of minima.

\section{Chaotic loss across a variety of domains}
In this section we will demonstrate chaotic dynamics which result in poorly behaved gradients in a variety of different systems. Note that these systems are not chaotic over the entire state space.
It's often possible to find restrictions to the problem that restore stability. For example, in a rigid body simulator, simulating in a region without contacts. Or in the case of optimization trajectories, simulating with an extremely small learning rate.  However, in many cases, these regions of state space are not the most ``interesting'', and running optimization most generally moves one towards regions of instability \citep{xiao2020disentangling}.
Chaotic dynamics in iterative systems have been demonstrated in physical systems in \citet{parmas2018pipps}, in optimization in \citep{pearlmutter1996investigation, maclaurin2015gradient} and learned optimization in \citep{metz2019understanding}.

\begin{figure}
    \centering
    \makebox[\textwidth]{%
    \begin{overpic}[width=0.4\textwidth]{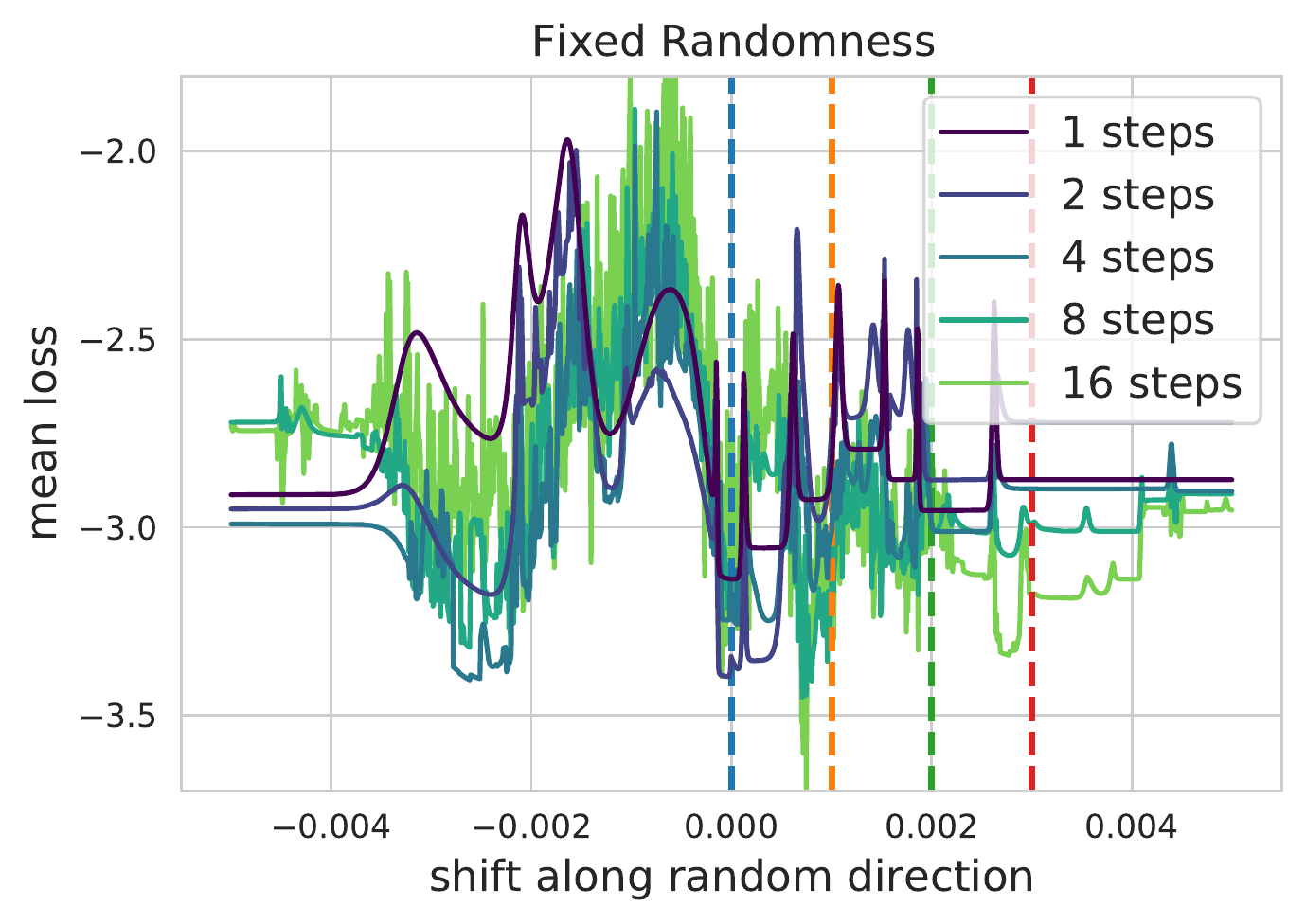}
    \put (0,70.0) {\textbf{\small(a)}}
    \end{overpic}
    \begin{overpic}[width=0.4\textwidth]{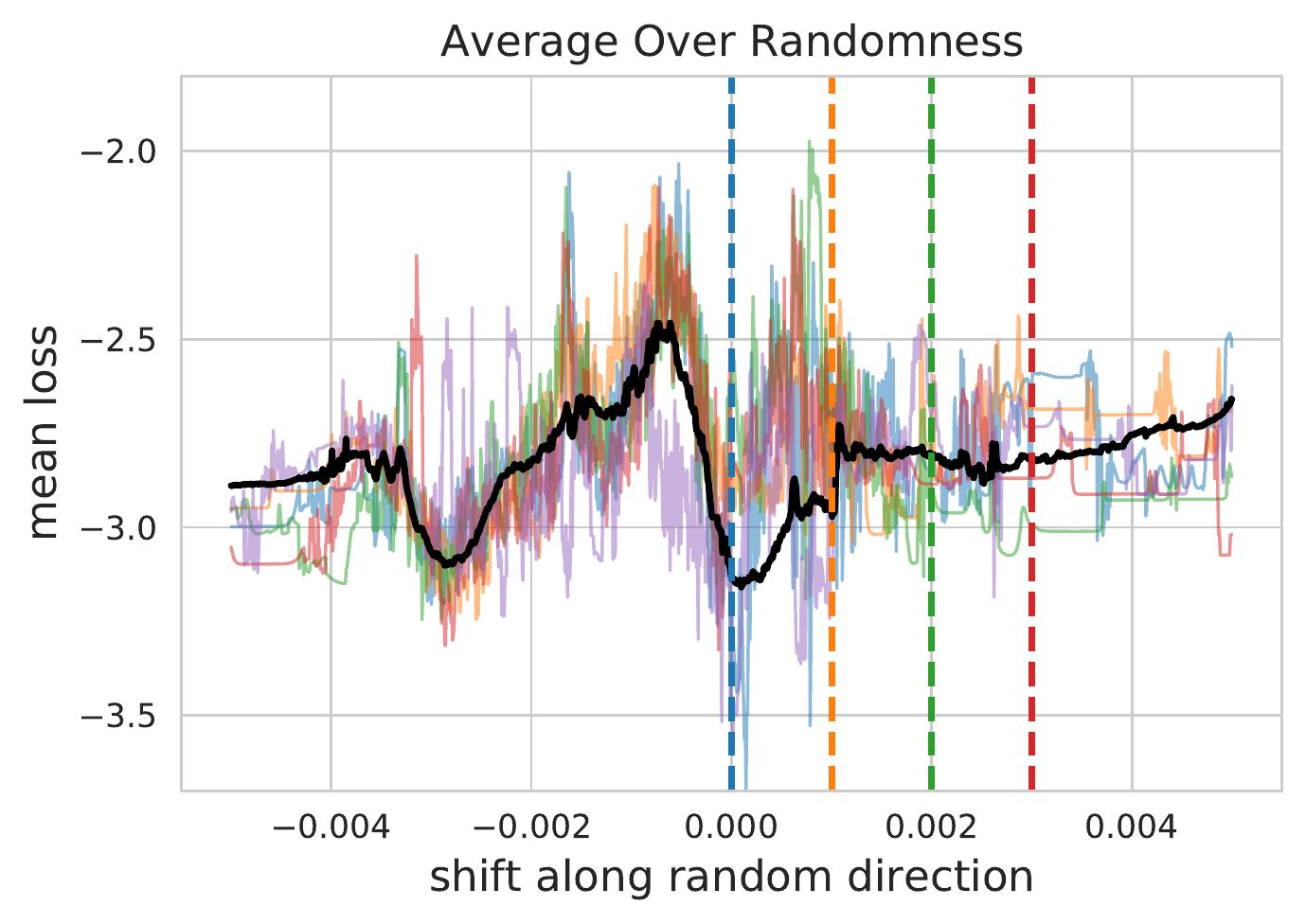}
    \put (0,70.0) {\textbf{\small(b)}}
    \end{overpic}
    \begin{overpic}[width=0.4\textwidth]{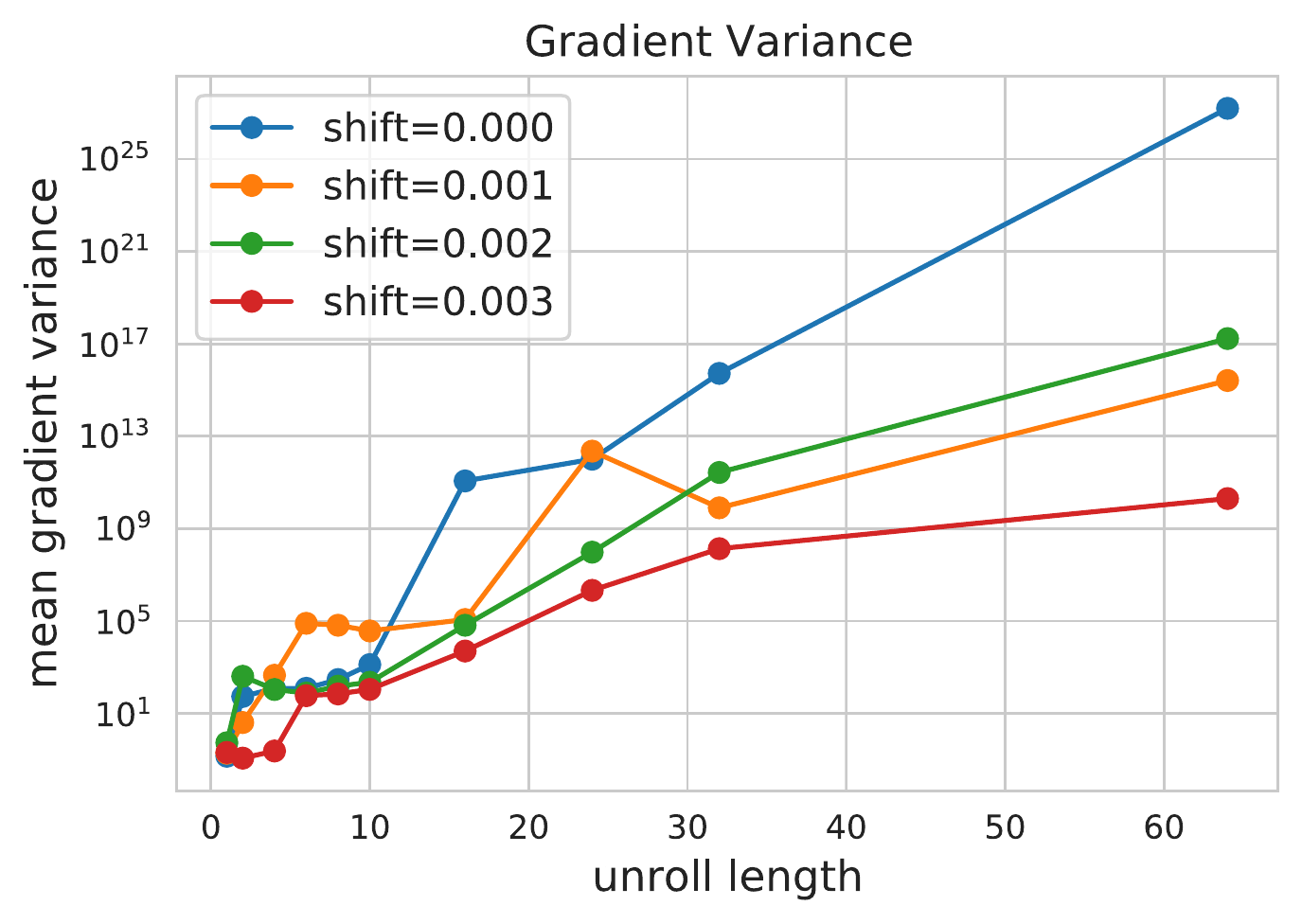}
    \put (0,70.0) {\textbf{\small(c)}}
    \end{overpic}
    }
    \vspace{-1em}
    \caption{Loss surface and gradient variance for a stochastic policy on a a robotics control task -- the Ant environment from Brax.
    \textbf{(a)}: We show a 1D projection of the loss surface along a random direction. All randomness in this plot is fixed. Color denotes different lengths of unroll when computing the loss. For small numbers of iterations the loss is smooth. For higher numbers of steps the underlying loss becomes highly curved. \textbf{(b)}: Instead of fixing randomness as done in the left plot, we average over multiple random samples for the 8 step unroll (average is in black, samples are in colors). We find that averaging greatly smooths the underlying loss surface. \textbf{(c)}: We look at gradient variance of gradients computed over multiple random samples from the stochastic policy. We show three different parameter values (shifts corresponding to the x-axis in the first two plots and are denoted with the same color vertical dashed lines). Despite having a seemingly smooth loss surface, the gradient variance explodes in exponential growth.
    \label{fig:ant}
    }
\end{figure}

\subsection{Rigid Body Physics}
First, we consider differentiating through physics simulations.
We make use of the recently introduced Brax \citep{freeman2021brax} physics package and look at policy optimization of some stochastic policy parameterized via a neural network. We test this using the default Ant environment and default MLP policy. For all computations, we use double precision floating point numbers.

For each environment, we first fix randomness and evaluate the loss along a fixed, random direction for different numbers of simulation steps (figure \ref{fig:ant}a).
For small number of steps, the resulting loss appears smooth, but as more and more steps are taken the loss surface becomes highly sensitive to the parameters of the dynamical system ($\theta$).
Next, we show the loss surface, with the same shifts, averaged over a number of random seeds(figure \ref{fig:ant}b). This randomness controls the sampling done by the policy. This averaged surface is considerably better behaved, similar to what was shown in \citet{parmas2018pipps}.
Finally, we look at the variance of the gradients computed over different random seeds as a function of the unroll length(figure \ref{fig:ant}c).
We test 4 different locations to compute gradients -- using the same shift direction used in the 1D loss slices. Despite the smoothed loss due to averaging over randomness, we find an exponential growth in gradient variance as unroll length increases and great sensitivity to where gradients are being computed.
When averaging over gradient samples we can reduce variance as $1/\sqrt{N}$ where $N$ is the number of samples, but this quickly computationally infeasible as gradients norm can grow exponentially!

\subsection{Meta-learning: Backpropagation through learned optimization}
Next we explore instabilities arising from backpropagating through unrolled optimization in a meta-learning context.
We take the per parameter, MLP based, learned optimizer architecture used in \citep{metz2019understanding} and use this to train 2 layer, 32 hidden unit MLP's on MNIST \citep{lecun1998mnist}.
To compute gradients with respect to the learned optimizer parameters, we iteratively apply the learned optimizer using inner-gradients computed on a fixed batch of data.
Analogous to the sum of rewards in the previous section, we use the average of the log loss over the entire unroll as our meta-objective -- or the objective we seek to optimize the weights of the learned optimizer against.

\begin{figure}
    \centering
    \makebox[\textwidth]{%
    \begin{overpic}[width=0.4\textwidth]{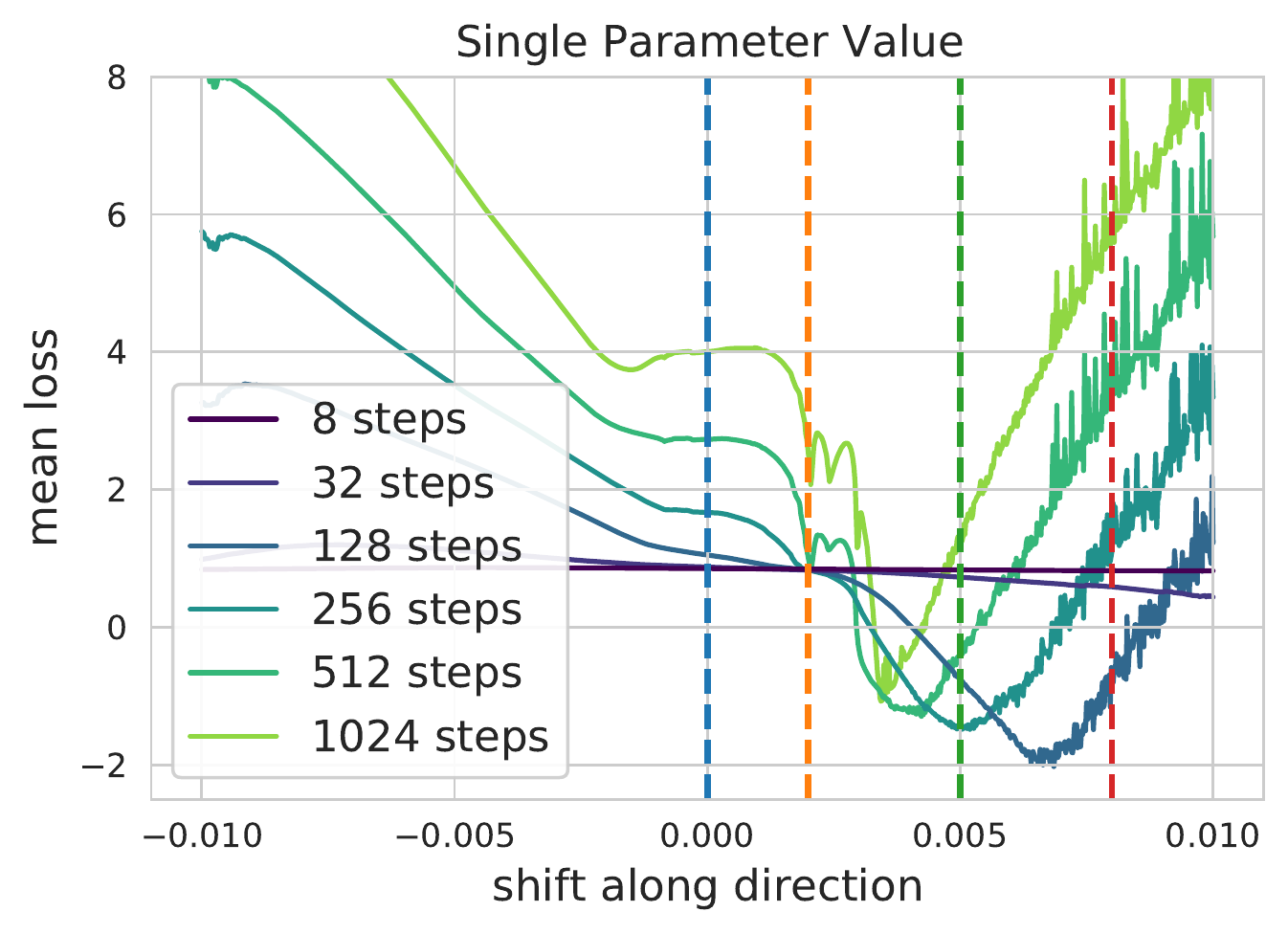}
    \put (0,70.0) {\textbf{\small(a)}}
    \end{overpic}
    \begin{overpic}[width=0.4\textwidth]{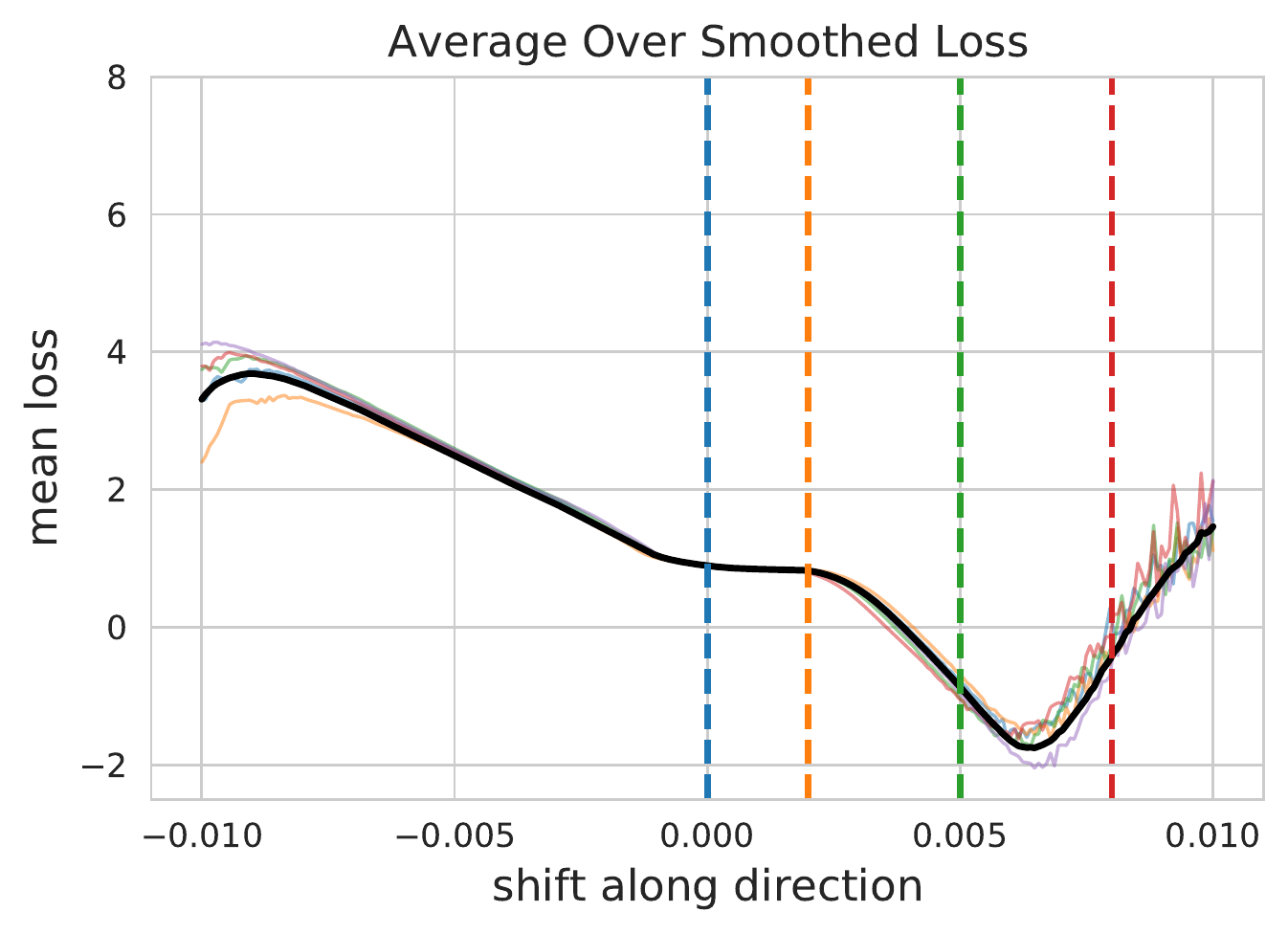}
    \put (0,70.0) {\textbf{\small(b)}}
    \end{overpic}
    \begin{overpic}[width=0.4\textwidth]{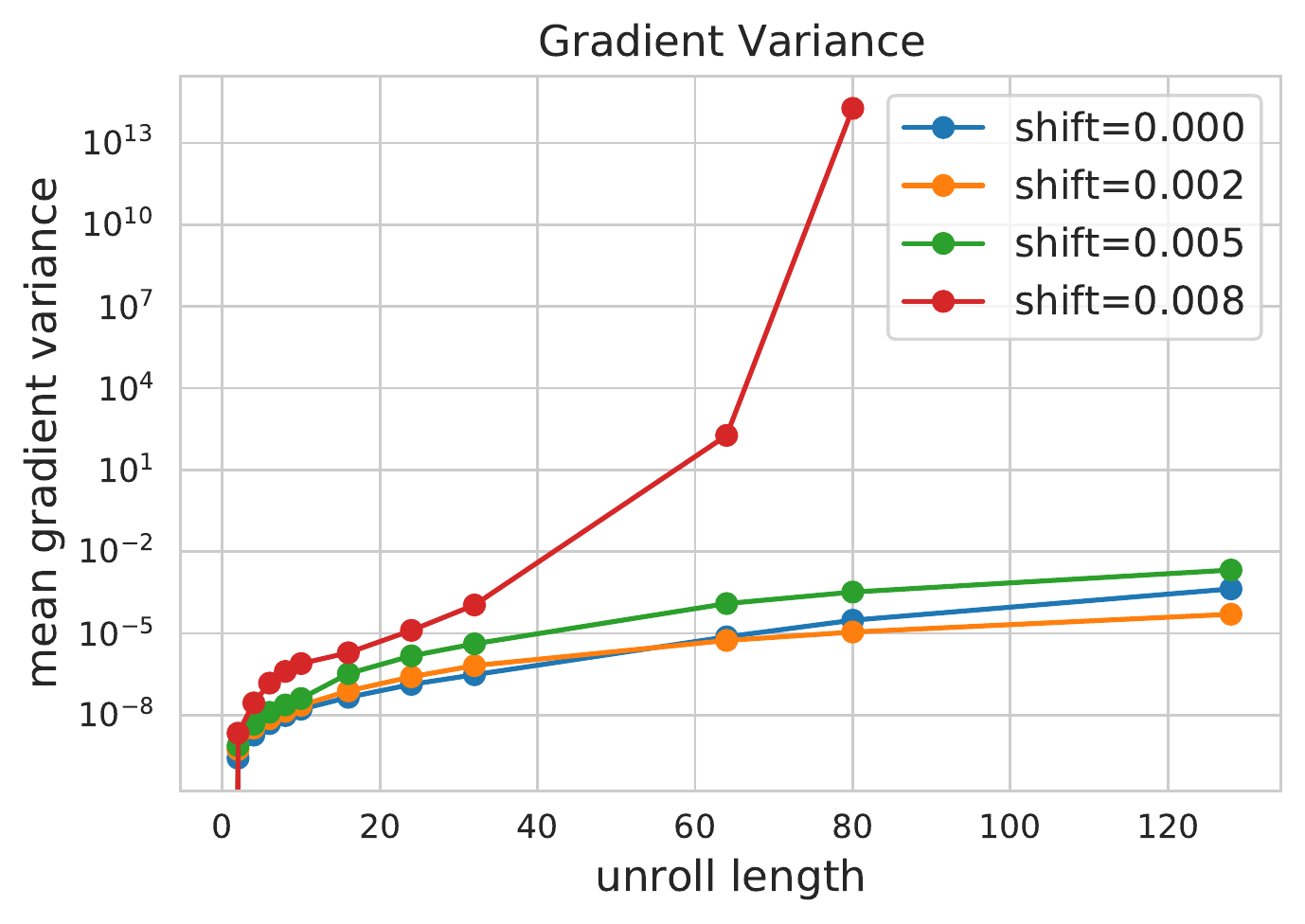}
    \put (0,70.0) {\textbf{\small(c)}}
    \end{overpic}
    }
    \vspace{-1em}
    \caption{Loss surface and gradient variance calculations for meta-learning an optimizer. \textbf{(a)}: We show a 1D projection of the meta-loss surfaces (loss with respect to learned optimizer parameters) for different length unrolls -- in this case, different numbers of application of the learned optimizer. For small numbers of steps, we find a smooth loss surfaces, but for higher numbers of steps we see a mix of smooth, and high curvature regions. \textbf{(b)}: We show an average of the meta-loss over Gaussian perturbed learned optimizer weights. The average is shown in black, and the losses averaged over are shown in color. We find this averaged loss is smooth and appears well behaved.
    \textbf{(c)}: We plot gradient variance over the different perturbations of the learned optimizer weights. These perturbations are shifts corresponding to the x-axis in the first two figures and are marked there with colored dashed vertical lines. For some settings of the learned optimizer weights (corresponding to the x-axis of the first 2 figures) we find well behaved gradient variance. For others, e.g. red, we find exponential growth in variance
    \label{fig:lopt}}
\end{figure}

In figure~\ref{fig:lopt}a, we show this loss computed with different length unrolls (shown in color).
We can see the same ``noisy'' loss surfaces with increased unroll length as before despite again having no sources of randomness.
Not all parameter values of learned optimizer parameter are sensitive to small changes in value.
Many randomly chosen directions resulted in flat, well behaved loss landscapes.
For this figure, we selected an initialization and direction out of 10 candidates to highlight this instability.
In figure~\ref{fig:lopt}b, we numerically compute the average loss smoothed around the current learned optimizer parameter value by a Gaussian with a standard deviation of 0.01 similar to what is done in \citep{metz2019understanding}.
We see that this smoothed loss surface appears to be well behaved -- namely low curvature.
Finally in figure~\ref{fig:lopt}c we compute the variance of the meta-gradient (gradient with respect to learned optimizer weights) over the loss smoothed by the normal distribution.
As before, we compute this gradient at different parameter values. For some parameter values the gradient variance grows modestly. In others, such as the 0.008 shifted value, the parameter space is firmly in the unstable regime and the gradient variance explodes.

\subsection{Molecular Dynamics}

Finally, we tune the properties of a simple material by differentiating through a molecular dynamics trajectory. In particular, we consider a variant of the widely studied ``packing problem''~\citep{LODI2002241} in which disks are randomly packed into a box in two-dimension~\citep{ohern2003}.
We take a bi-disperse system composed of disks of two different diameters in a box of side-length $L$; the smaller disks are taken to have diameter $D$, while the larger disks have a fixed diameter of one.
We fix the side-length so that the volume of space taken up by the disks, called the packing fraction, is constant (set to $\phi=0.98$) as we vary $D$.
It is well-known that different choices of $D$ lead to significantly different material properties: when $D$ is close to one the system forms a hexagonal crystal that packs nicely into the space; when $D\ll1$ the smaller disks fit into the interstices of the larger disks that once again form a hexagonal crystal; however, when $D\sim0.8$ the packing becomes disordered and the disks are not able to pack as tightly.
To generate packings, we use JAX MD~\citep{jaxmd2020} to produce some initial configuration of disks and then use a momentum-based optimizer called FIRE~\citep{bitzek2006structural} to quench the configuration to the nearest minimum. 

Previous work showed that $D$ could be tuned to find the maximally disordered point by differentiating through optimization~\citep{jaxmd2020} provided the system was initialized close to a stable packing.
In this regime, the optimization procedure was not chaotic since small changes to the initial configuration of disks will lead to the same final packing.
On the other hand, if the disks are randomly initialized then the dynamics become chaotic, since small changes to the initial configuration will lead to significantly different packings.
To demonstrate that these chaotic dynamics spoil gradient estimates computed using automatic differentiation, we randomly initialize disks varying $D$ and the random seed; we compute the derivative of the final energy with respect to $D$ by differentiating through optimization.

\begin{figure}
    \centering
    \makebox[\textwidth]{%
    \begin{overpic}[width=1.0\textwidth]{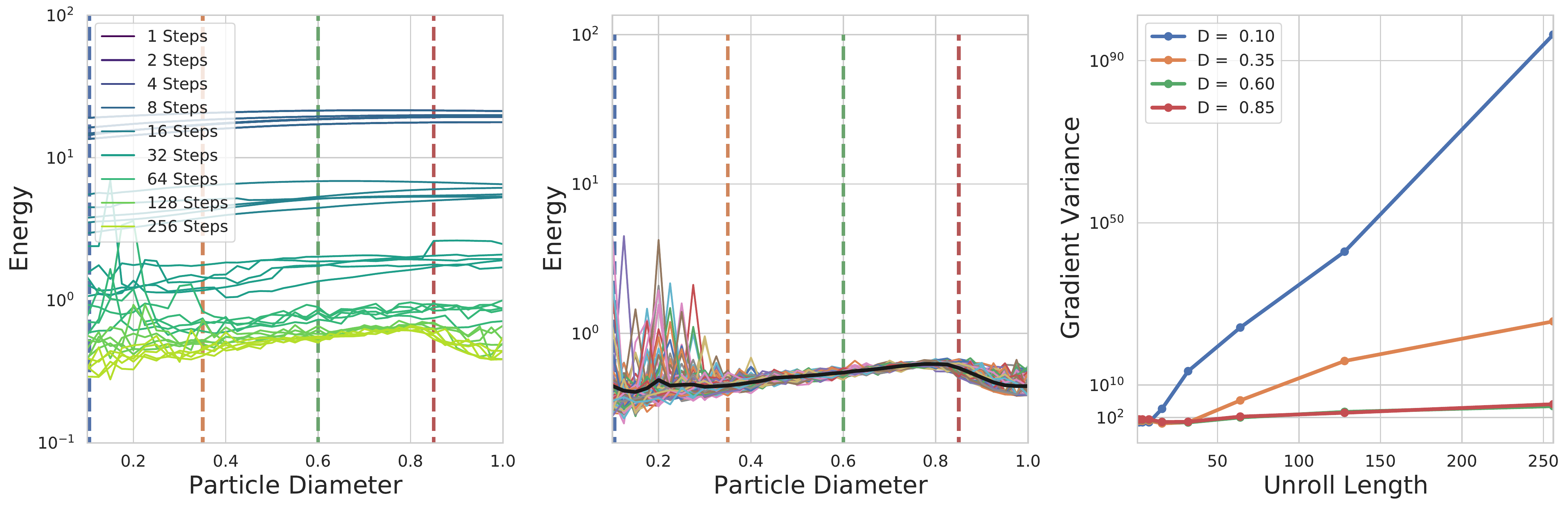}
    \put (2,33.0) {\textbf{\small(a)}}
    \put (36,33.0) {\textbf{\small(b)}}
    \put (68,33.0) {\textbf{\small(c)}}
    \end{overpic}
    }
    \vspace{-1em}
    \caption{Energy for packings of bi-disperse disks varying the diameter of the small disk, $D$, and the number of optimization steps. \textbf{(a)}: The energy of the system as a function of $D$ for different numbers of optimization steps. We see that the energy decreases with more steps of optimization. \textbf{(b)}: The energy for the maximum number of optimization steps considered (256). Each individual curve is the energy for one random configuration and the black line indicates the energy averaged over many random seeds.
    \textbf{(c)}: The variance of the gradient estimate for different values of $D$ as a function of the number of steps of optimization.}
    \label{fig:jax_md}
\end{figure}

In figure~\ref{fig:jax_md} we show the results of differentiating the energy through optimization with respect to the diameter.
In figure~\ref{fig:jax_md}a we see the energy for a number of different random seeds. As the number of steps of optimization grows, the energy decreases but the variance across seeds increases.
In figure~\ref{fig:jax_md}b we see the energy for a number of different random seeds after 256 steps of optimization along with the energy averaged over seeds.
We see that, especially for small diameters, the variance is extremely large although the average energy is well-behaved.
Finally in figure~\ref{fig:jax_md}c, we see the variance of the gradients as a function of optimization steps for several different diameters.
We see that the variance grows quickly, especially for the smallest particle diameters.

\subsection{Connecting gradient explosion to spectrum of the recurrent Jacobian}
\begin{figure}
    \centering
    \makebox[\textwidth]{%
    \begin{overpic}[width=1.2\textwidth]{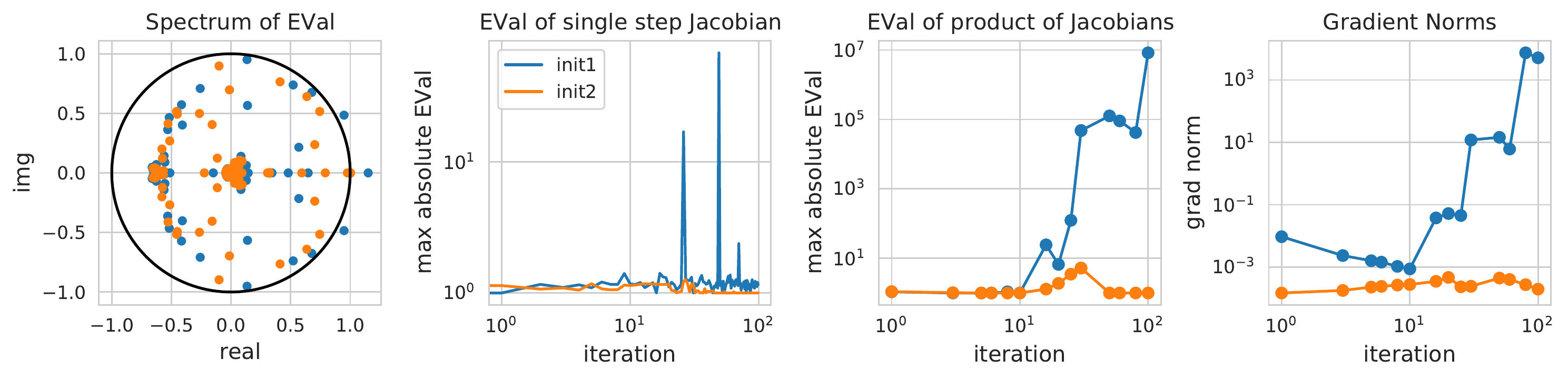}
    \put (2,24.0) {\textbf{\small(a)}}
    \put (27,24.0) {\textbf{\small(b)}}
    \put (53,24.0) {\textbf{\small(c)}}
    \put (79,24.0) {\textbf{\small(d)}}
    \end{overpic}
    }
    \vspace{-1em}
    \caption{Exploration into the eigenspectrum of the recurrent jacobians of the Brax Ant experiment. We show two parameter values: init1 which is initialized in an unstable regime, and init2 which is in stable regime. \textbf{(a)}: we show the spectrum of the recurrent jacobain taken from the 90th iteration ($\pd{s_{90}}{s_{89}}$). \textbf{(b)}: We plot the length of the maximum norm eigenvalue of each recurrent jacobian along the sequence ( $\pd{s_{i}}{s_{i-1}}$ for each $i$). \textbf{(c)}: We plot the length of the max eigenvalue of the ($\pd{s_{i}}{s_{0}}$ for each $i$) and find that the unstable initialization grows exponentially.
    \textbf{(d)}: We plot the gradient norms of each initialization and find exploding gradients in the unstable initialization.
    \label{fig:jac}
    }
\end{figure}
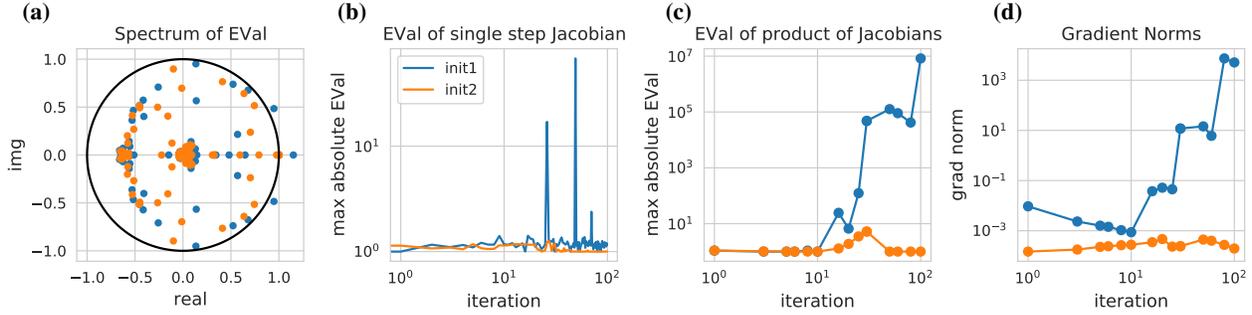

To better understand what causes these gradient explosions, we look to measuring statistics of the recurrent Jacobian ($\pd{s_i}{s_{i-1}}$) as well as the product of recurrent Jacobians ($\prod_{i=0}^{t} \pd{s_i}{s_{i-1}}$).
We do this experimentally on the Ant environment with two different parameter values with which to measure at.
First, a random initialization of the NN policy (init1), which is poorly behaved and results in exploding gradient norms, and an initialization which shrinks this initialization by multiplying by 0.01. This second initialization was picked so that gradient would not explode. With these two initializations, we plot the spectrum of the Jacobian, the max eigenvalue of the Jacobian for each iteration (i.e. $\pd{s_i}{s_{i-1}}$ for all $i$), the cumultive max absolute of the product of jacobians (i.e. $\pd{s_i}{s_{0}}$ for all $i$), and the gradient norm for a given unroll length. Results in figure \ref{fig:jac}.

For the unstable initialization (init1) we find many eigenvalues with norm greater than length 1 (figure~\ref{fig:jac}ab blue), and thus find the cumulative max eigen value grows exponentially (figure~\ref{fig:jac}c blue) and thus gradient norms grow(figure~\ref{fig:jac}d blue). For the stable initialization (init2), we find many eigenvalues close to 1 (figure~\ref{fig:jac}ab orange), resulting in little to no growth in the max eigenvalue (figure~\ref{fig:jac}c orange) and thus controlled gradient norms(figure~\ref{fig:jac}d orange).

\section{What can be done?}
If we cannot naively use gradients to optimize these systems what can we use? In this section we discuss a couple of options often explored by existing work.

\subsection{Pick well behaved systems}
If exploding gradients emerge from chaotic dynamics of the underlying system, one way to ``solve'' this problems is to change systems.
Whether or not this is really a ``solution'' is, perhaps, a matter of perspective. For example, empirically it seems that many molecular physics systems naturally have dynamics that are not particularly chaotic and gradient-based optimization can be successfully employed (for example~\citep{goodrich2021designing, kaymak2021jax}).
For many types of systems, e.g. language modeling, the end goal is to find a high performance model, so it doesn't matter if one particular type of architecture has difficult to optimize, chaotic dynamics---we're free to simply pick a more easily optimize-able architecture.
We discuss these modifications and why they work in \ref{sec:rnn}. 
For other systems, such as rigid-body physics, changing the system in this way will be biased, and such bias could effect one's downstream performance.
For example, in rigid-body physics simulations we want to simulate physics, not a non-chaotic version of physics.
We discuss modifications that can be done in \ref{sec:physics}.

\subsubsection{Recurrent neural networks} \label{sec:rnn}
One general dynamical system explored in deep learning which also has these types of exploding or vanishing gradients are recurrent neural networks (RNN).
The gradient of a vanilla RNN exhibits exactly the same exponentially sensitive dynamics described by Eq. \ref{eq:product}, with vanishing/exploding gradients depending on the jacobian of the hidden state parameters.\citep{pascanu2013difficulty}.

Of the many solutions discussed, we will highlight 2 which overcome this issue: different initializations, and different recurrent structure.

\textbf{Change the initialization:} IRNN\citep{le2015simple} work around this problem by initializing the RNN near the identity. At initialization this means the recurrent Jacobian will have eigenvalues near 1 and thus be able to be unrolled longer before encountering issues.
After training progresses and weights update, however, the Jacobian drifts, eventually resulting in vanishing/exploding gradients late enough in training.

\textbf{Change recurrent structure:} A second solution is to change the problem entirely. In the case of an RNN this is feasible by simply changing the neural architecture. LSTM's \citep{hochreiter1997long}, GRU \citep{chung2014empirical}, UGRNN \citep{collins2016capacity} are such modifications.
As shown in \citep{bayer2015learning}, the recurrent jacobian of an LSTM was specifically designed to avoid this exponential sensitivity to the hidden state, and is thus significantly more robust than a vanilla RNN.
While changing architecture does increase stability, it comes at the cost of no-longer being able to model chaotic relationships. \citet{monfared2021train} discuss this, and the relationship chaos and the gradients of the loss.

\subsubsection{Rigid Body Physics} \label{sec:physics}
Physics simulation involving contact is notoriously difficult to treat differentiably.
The problem arises from sharp changes in object velocity before and after a contact occurs (e.g., a ball bouncing off of a wall). Various methods have been developed recently to circumvent this issue.
Contact ``softening'' has proven particularly fruitful, where contact forces are blurred over a characteristic lengthscale, instead of being enforced at a sharp boundary \citep{huang2021plasticinelab}.
Others have explicitly chunked the process of trajectory optimization into a sequence of mini-optimizations demarcated by moments of contact \citep{cleach2021fast}.

\begin{figure}
    \centering
    \vspace{-1em}
    \includegraphics[width=0.5\textwidth]{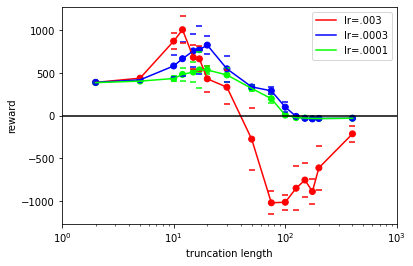}
    \caption{Reward on a modified version of the ant locomotion task in Brax.  In this task, we backpropagate the task reward directly to the policy parameters after 400 steps in the environment.  For truncation length $t$, a stop\_gradient op was inserted every $t$ steps of the 400 step trajectory.  Short truncations typically optimize a lunging policy that results in the ant moving a short distance to the right, but staying on its feet.  Long truncations uniformly fail to optimize.  Truncations around length $10$ result in successful locomotive policies.    \label{fig:ant_truncations}}

\end{figure}

While these methods have proven fruitful in several domains, they introduce significant algorithmic complexity.
This may be a necessary cost---i.e., difficult problems requiring difficult solutions---but we have also found black box methods to extremely reliably solve these problems without needing to introduce any task-specific algorithmic tuning (Section \ref{sec:blackbox}).
It's surprising, then, that gradients seem to \emph{introduce} complexity to the task of trajectory optimization in robotics.

\subsubsection{Well-behaved proxy objectives}

In some cases, especially in statistical physics systems, features of the energy landscape that govern properties of interest are known. These features can serve as proxy objectives that can be optimized using automatic differentiation without differentiating through long simulation trajectories. In atomistic systems, for example: the eigenvalues of the hessian near minima -- called phonons -- control properties ranging from heat transport to stiffness~\citep{ashcroft1976solid}, the height of saddle points in the landscape compared to the minima control the rate at which the system moves between minima~\citep{truhlar1996current}. These properties have been successfully optimized using automatic differentiation~\citep{jaxmd2020, blondel2021efficient, goodrich2021designing} and do not suffer from bias as in the case of truncated gradients. Of course, this approach can only be exploited when such proxy objectives are known \textit{a priori}. 

In cases where the property of the landscape that we are trying to optimize is known and can be phrased in terms of the fixed point of some dynamical system, there can often be significant benefit to computing derivatives implicitly rather than unrolling optimization explicitly~\cite{rajeswaran2019meta, bai2019deep, blondel2021efficient}. Using implicit differentiation in this way  removes conditioning issues due to unrolling and improves memory cost of computing derivatives. However, these methods are only well-defined when the iterated map converges deterministically to a single fixed point. In the case of gradient descent, this often corresponds to initializing the solver in a convex region around minima of the landscape. As such, implicit gradients do not fundamentally resolve the issue of chaos in the general case.

It is also possible to modify the underlying loss surface to better regularize it. \citet{ingraham2018learning} regularize their dynamics function to be approximarly 1-Lipshitz. If exact this would imply chaos is impossible. In climate sciences, ``nudging'' has been proposed \citep{abarbanel2010data, kohl2002adjoint}. These methods modify the underlying loss with a term that ``nudges'' the state to stay near some reference trajectory. By regularizing this way, gradients over many iterations grow at a slower rate. While they have been used to tune parameters of weather simulators, we are not aware of any applications in machine learning.

\subsection{Truncated backpropogation}
Another common technique used to control these issues is called truncated backpropogation through time \citep{werbos1990backpropagation, tallec2017unbiasing}.
This has been used to great effect in training of language models \citep{sutskever2013training}, learned optimizers \citep{andrychowicz2016learning, wichrowska2017learned, lv2017learning, metz2019understanding}, and fluid simulation \citep{kochkov2021machine}.
This type of solution comes at the cost of bias, however, as the gradients computed are missing terms -- in particular, the longer products of Jacobians from equation \ref{eq:product}.
Depending on the underlying task this bias can be severe \citep{wuunderstanding}, while for other tasks, empirically, this matters less.

To demonstrate the effects of truncation we attempt to train a policy for the Ant environment in Brax using batched stochastic gradient descent, visualized in Fig. \ref{fig:ant_truncations}.
Learning via the raw gradient signal backpropagated through 400 environment steps catastrophically fails to find any useful policy despite an extensive hyper parameter search.
However, with truncated gradient updates, there exists a narrow band of truncation lengths at which the Ant policy is able to successfully learn a locomotive gait.

Truncated backpropogation operates by stopping backpropogated signal. \citet{ingraham2018learning} soften this approximation by decaying the gradient signal each iteration and thus lowering the eigenvalues of the recurrent Jacobian.

\subsection{Gradient clipping}
Another common technique to train in the precesses of exploding gradients is gradient clipping \citet{pascanu2013difficulty}. Instead of taking the true gradients, one can train using gradients clipped in some way. This has proven to be of use in a variety of domains\citep{kaiser2015neural, merity2017regularizing, gehring2017convolutional} but will not fix all problems. As before, this calculation of the gradient is biased. To demonstrate this, we took the same Ant policy and sweep learning rate and gradient clipping strength. We found no setting which results in positive performance and thus omitted the plot.

\subsection{Methods for ergodic systems}
In some cases, the underlying dynamical system is ergodic which enable more sophisticated gradient estimation techniques. Loosely speaking, ergodicity here means this means that over sufficiently long times, the dynamical system will uniformly sample from a distribution over states in a time-independent manner. Examples of this often arise in physical systems composed of many constituents such as fluids, materials, or models of the climate.

\subsubsection{Least Squares Shadowing}
Shadowing methods tackle the problem of computing derivatives through infinite length time averages. 
To do this they take advantage of the shadowing lemma\citep{pilyugin2006shadowing}. The shadowing lemma tells us that for every trajectory computed with some numerical error (say rounding errors) stays uniformly close to some true trajectory.
Least squares shadowing (LSS) methods make use of this to find a better behaved nearby trajectory and then use this trajectory to compute gradients. To our knowledge, these methods are most often discussed in the continuous time regime requiring differential equation solvers instead of discrete simulation steps, and tangent solvers instead of simple back propogation. 

This idea was first introduced in \citep{wang2014least} with further which construct a least squares problem to find these nearby trajectories, linearize then solves it and uses this solution trajectory to estimate gradients. Non-intrusive LSS (NILSS) \citep{ni2017sensitivity} extend this by reducing computation only to the unstable subspace and Finite-Difference NILSS (FD-NILSS) \citep{ni2019sensitivity} leverages finite difference as opposed to a tangent solver. For an in-depth discussion see \citet{ni2021numerical}.
These methods were applied to sensitivity analysis for fluids in turbulent systems \citet{blonigan2014least} and large-eddy simulation \citep{blonigan2017toward}.

\subsubsection{Inverting the Shadow Operator}
\citet{wang2013forward} propose another approach to compute derivatives through chaotic systems which also leverages the shadowing lemma and assumes infinite time averages. A ``Shadowing Operator'' is constructed which which maps between perturbed and true coordinates and inverted to compute gradients. \citet{wang2013forward} include both a forward mode differentiation and reverse mode (adjoint method).

\subsubsection{Probabilistic approaches}
In these approaches instead of working with samples, one works with the distribution of states directly. This idea was explored in climate modeling leveraging the Fokker-Planck equation \citep{thuburn2005climate}. Other methods make use of the Fluctuation-Dissipation Theorem with various diferent assumptions \citep{abramov2007blended, cooper2011climate}. \citet{gutierrez2019response} cast the evolution of the state space distribution as Markov Chains enabling gradient calculation. \citet{chandramoorthy2020computable} 

Another family of methods to compute time averaged gradient involve computing averages over an ensemble of finite horizon evaluations. This concept was first proposed in \citep{lea2000sensitivity} and improved upon by the space-split sensitivity algorithm (S3) \citep{chandramoorthy2021efficient} to have better convergence rates. See \citep{chandramoorthy21thesis} for a more in depth discussion of these, and other methods for differentiating through ergodic systems.

\subsection{Learned Models}
While differentiating through chaotic dynamics is challenging, it is often much simpler to learn to approximate these functions, and then use the approximation for some other task. Learning models like this is commonplace across many domains:  fluid simulation \citep{ladicky2015data, tompson2017accelerating, kochkov2021machine}, control \citep{ha2018world}, and reinforcement learning of games\citep{hafner2019learning, ozair2021vector}. These methods ``collapse'' an otherwise complex iterative process with a more shallow computation (e.g. a forward pass through a neural network). In addition to being better behaved, these approximate function are often faster to compute than the original simulation. These approximate functions can then be used in place of, or used in combination with the original dynamics.

To our knowledge, the main use of this family of method is in model based reinforcement learning where one learns a model of an environment and uses this model for control or planning.
Instead of differentiating through this learned model, however, many methods instead learn a value function \footnote{or Q function} which determines performance in the future then differentiates though a handful of steps before truncating the trajectory with this learned value \citep{feinberg2018model, buckman2018sample,kurutach2018model, clavera2020model}.

\subsection{Just use black box gradients} \label{sec:blackbox}
One somewhat naive sounding solution is to throw away all gradients and resort a some black box method to estimate gradients. In some cases this also forms an unbiased estimate of the exact same gradient of interest. For example REINFORCE with no discounting \cite{williams1992simple} can be used to estimate the same gradient as computed in the Brax experiments, and evolutionary strategies \citep{rechenberg1973evolutionsstrategie, schwefel1977evolutionsstrategien, wierstra2008natural, salimans2017evolution} can be used to estimate the gradient for the the learned optimizers.

By resorting to a black box method, we theoretically lose a factor of dimensionality of efficiency when estimating gradients. In practice, however, and given that gradient variance can grow exponentially, computing gradients in this way can lead to lower variance estimates. We demonstrated a sketch of this in figure~\ref{fig:reparm_es}. Instead of picking one or the other estimator, one can instead combine the two gradient estimates to produce an even lower variance estimate of the gradients. In the context of continuous control this has been explored in depth as a solution to high variance gradients arising from chaos in \citet{parmas2018pipps}, and in the context of learned optimizers in \citet{metz2019understanding}.

In addition to unbiased methods, there are a host of other methods with varying bias/variance properties that can also be used -- most of which coming from the Deep RL community. For example, PPO~\citep{schulman2017proximal} easily outperforms all of our experiments training the Ant policy with gradients we performed.

\section{Discussion}
In this paper we dive into chaos as a potential issue when computing gradients through dynamical systems.
We believe this is one particularly sinister issue that is often ignored and hence the focus of this work.
Many other potential issues also exist.
Numerical precision for these unrolled systems is known to be a problem and addressed to some extent by \citet{maclaurin2015gradient}. Memory requirements are often cited as an issue for backprop as naively it requires $N$ times as much memory as a forward pass where $N$ is the length of the sequence, or $\sqrt{N}$ memory with gradient checkpointing\citep{griewank2000algorithm, chen2016training}.
Finally, many systems of interest could have extremely flat, or hard to explore loss landscapes making gradient based learning extremely tricky.
Part of the reason gradient descent works in neural networks is due to over parameterization \citep{kawaguchi2016deep} and known weights prior/initialization, which is often not possible in simulated systems have.

Despite the large number of issues computing gradients through recurrent processes, they have shown many wonderful results! We hope this paper sheds light into when gradients can be used -- when the recurrent Jacobian has small eigenvalues. In the other cases, when gradients do not work, we encourage readers to try black box methods -- they estimate the same quantity and with less pathological variance properties, especially when it's possible to calculate a smoothed proxy for the loss function of interest.
In summary, gradients are not all you need. Just because you can take a gradient doesn't mean you always should.

\begin{ack}
We thank Stephan Hoyer, Jascha Sohl-Dickstein, and Vincent Vanhoucke for feedback on early drafts of this manuscript, as well as the rest of the Google Brain and Accelerated Science teams for their support. We would also like to thank
Jonathan Balloch,
Noah Brenowitz, 
Thomas Bingel,
Daniel Durstewitz, 
Joshua Kimrey,
Andrea Panizza,
Paavo Parma,
Ludger Paehler,
Chris Rackauckas,
and Molei Tao for their tweets and emails which helped us improved the first version of this work.
T.K would like to acknowledge Lineage Logistics for hosting and partial funding while this research was performed. T.K also thank Kell's establishment and Elliot Wolf.
\end{ack}

\clearpage

\bibliography{references}

\begin{thebibliography}{111}
\providecommand{\natexlab}[1]{#1}
\providecommand{\url}[1]{\texttt{#1}}
\expandafter\ifx\csname urlstyle\endcsname\relax
  \providecommand{\doi}[1]{doi: #1}\else
  \providecommand{\doi}{doi: \begingroup \urlstyle{rm}\Url}\fi

\bibitem[Heiden et~al.(2021)Heiden, Millard, Coumans, Sheng, and
  Sukhatme]{heiden2021neuralsim}
Eric Heiden, David Millard, Erwin Coumans, Yizhou Sheng, and Gaurav~S Sukhatme.
\newblock Neural{S}im: Augmenting differentiable simulators with neural
  networks.
\newblock In \emph{Proceedings of the IEEE International Conference on Robotics
  and Automation (ICRA)}, 2021.
\newblock URL
  \url{https://github.com/google-research/tiny-differentiable-simulator}.

\bibitem[Hu et~al.(2019)Hu, Li, Anderson, Ragan-Kelley, and
  Durand]{hu2019taichi}
Yuanming Hu, Tzu-Mao Li, Luke Anderson, Jonathan Ragan-Kelley, and Fr{\'e}do
  Durand.
\newblock Taichi: a language for high-performance computation on spatially
  sparse data structures.
\newblock \emph{ACM Transactions on Graphics (TOG)}, 38\penalty0 (6):\penalty0
  201, 2019.

\bibitem[Werling et~al.(2021)Werling, Omens, Lee, Exarchos, and
  Liu]{werling2021fast}
Keenon Werling, Dalton Omens, Jeongseok Lee, Ioannis Exarchos, and C~Karen Liu.
\newblock Fast and feature-complete differentiable physics for articulated
  rigid bodies with contact.
\newblock \emph{arXiv preprint arXiv:2103.16021}, 2021.

\bibitem[Degrave et~al.(2019)Degrave, Hermans, Dambre,
  et~al.]{degrave2019differentiable}
Jonas Degrave, Michiel Hermans, Joni Dambre, et~al.
\newblock A differentiable physics engine for deep learning in robotics.
\newblock \emph{Frontiers in neurorobotics}, 13:\penalty0 6, 2019.

\bibitem[de~Avila Belbute-Peres et~al.(2018)de~Avila Belbute-Peres, Smith,
  Allen, Tenenbaum, and Kolter]{de2018end}
Filipe de~Avila Belbute-Peres, Kevin Smith, Kelsey Allen, Josh Tenenbaum, and
  J~Zico Kolter.
\newblock End-to-end differentiable physics for learning and control.
\newblock \emph{Advances in neural information processing systems},
  31:\penalty0 7178--7189, 2018.

\bibitem[Gradu et~al.(2021)Gradu, Hallman, Suo, Yu, Agarwal, Ghai, Singh,
  Zhang, Majumdar, and Hazan]{gradu2021deluca}
Paula Gradu, John Hallman, Daniel Suo, Alex Yu, Naman Agarwal, Udaya Ghai,
  Karan Singh, Cyril Zhang, Anirudha Majumdar, and Elad Hazan.
\newblock Deluca--a differentiable control library: Environments, methods, and
  benchmarking.
\newblock \emph{arXiv preprint arXiv:2102.09968}, 2021.

\bibitem[Freeman et~al.(2021)Freeman, Frey, Raichuk, Girgin, Mordatch, and
  Bachem]{freeman2021brax}
C~Daniel Freeman, Erik Frey, Anton Raichuk, Sertan Girgin, Igor Mordatch, and
  Olivier Bachem.
\newblock Brax-a differentiable physics engine for large scale rigid body
  simulation.
\newblock 2021.

\bibitem[Li et~al.(2018)Li, Aittala, Durand, and
  Lehtinen]{li2018differentiable}
Tzu-Mao Li, Miika Aittala, Fr{\'e}do Durand, and Jaakko Lehtinen.
\newblock Differentiable monte carlo ray tracing through edge sampling.
\newblock \emph{ACM Transactions on Graphics (TOG)}, 37\penalty0 (6):\penalty0
  1--11, 2018.

\bibitem[Kato et~al.(2020)Kato, Beker, Morariu, Ando, Matsuoka, Kehl, and
  Gaidon]{kato2020differentiable}
Hiroharu Kato, Deniz Beker, Mihai Morariu, Takahiro Ando, Toru Matsuoka, Wadim
  Kehl, and Adrien Gaidon.
\newblock Differentiable rendering: A survey.
\newblock \emph{arXiv preprint arXiv:2006.12057}, 2020.

\bibitem[Schoenholz and Cubuk(2020)]{jaxmd2020}
Samuel~S. Schoenholz and Ekin~D. Cubuk.
\newblock Jax m.d. a framework for differentiable physics.
\newblock In \emph{Advances in Neural Information Processing Systems},
  volume~33. Curran Associates, Inc., 2020.
\newblock URL
  \url{https://papers.nips.cc/paper/2020/file/83d3d4b6c9579515e1679aca8cbc8033-Paper.pdf}.

\bibitem[Hinsen(2000)]{hinsen2000molecular}
Konrad Hinsen.
\newblock The molecular modeling toolkit: a new approach to molecular
  simulations.
\newblock \emph{Journal of Computational Chemistry}, 21\penalty0 (2):\penalty0
  79--85, 2000.

\bibitem[Maclaurin(2016)]{maclaurin2016modeling}
Dougal Maclaurin.
\newblock \emph{Modeling, inference and optimization with composable
  differentiable procedures}.
\newblock PhD thesis, 2016.

\bibitem[Bischof et~al.(1996)Bischof, Pusch, and
  Knoesel]{bischof1996sensitivity}
Christian~H Bischof, Gordon~D Pusch, and Ralf Knoesel.
\newblock Sensitivity analysis of the mm5 weather model using automatic
  differentiation.
\newblock \emph{Computers in Physics}, 10\penalty0 (6):\penalty0 605--612,
  1996.

\bibitem[McGreivy et~al.(2021)McGreivy, Hudson, and Zhu]{mcgreivy2021optimized}
Nick McGreivy, Stuart~R Hudson, and Caoxiang Zhu.
\newblock Optimized finite-build stellarator coils using automatic
  differentiation.
\newblock \emph{Nuclear Fusion}, 61\penalty0 (2):\penalty0 026020, 2021.

\bibitem[Paszke et~al.(2017)Paszke, Gross, Chintala, Chanan, Yang, DeVito, Lin,
  Desmaison, Antiga, and Lerer]{paszke2017automatic}
Adam Paszke, Sam Gross, Soumith Chintala, Gregory Chanan, Edward Yang, Zachary
  DeVito, Zeming Lin, Alban Desmaison, Luca Antiga, and Adam Lerer.
\newblock Automatic differentiation in pytorch.
\newblock In \emph{NIPS-W}, 2017.

\bibitem[Ablin et~al.(2020)Ablin, Peyr{\'e}, and Moreau]{ablin2020super}
Pierre Ablin, Gabriel Peyr{\'e}, and Thomas Moreau.
\newblock Super-efficiency of automatic differentiation for functions defined
  as a minimum.
\newblock In \emph{International Conference on Machine Learning}, pages 32--41.
  PMLR, 2020.

\bibitem[Margossian(2019)]{margossian2019review}
Charles~C Margossian.
\newblock A review of automatic differentiation and its efficient
  implementation.
\newblock \emph{Wiley interdisciplinary reviews: data mining and knowledge
  discovery}, 9\penalty0 (4):\penalty0 e1305, 2019.

\bibitem[Bischof et~al.(1991)Bischof, Griewank, and
  Juedes]{bischof1991exploiting}
Christian Bischof, Andreas Griewank, and David Juedes.
\newblock Exploiting parallelism in automatic differentiation.
\newblock In \emph{Proceedings of the 5th international conference on
  Supercomputing}, pages 146--153, 1991.

\bibitem[Corliss et~al.(2013)Corliss, Faure, Griewank, Hascoet, and
  Naumann]{corliss2013automatic}
George Corliss, Christele Faure, Andreas Griewank, Laurent Hascoet, and Uwe
  Naumann.
\newblock \emph{Automatic differentiation of algorithms: from simulation to
  optimization}.
\newblock Springer Science \& Business Media, 2013.

\bibitem[Chow and Palmer(1992)]{chow1992numerical}
Shui-Nee Chow and Kenneth~J Palmer.
\newblock On the numerical computation of orbits of dynamical systems: the
  higher dimensional case.
\newblock \emph{Journal of Complexity}, 8\penalty0 (4):\penalty0 398--423,
  1992.

\bibitem[Kachman et~al.(2017)Kachman, Fishman, and
  Soffer]{kachman2017numerical}
Tal Kachman, Shmuel Fishman, and Avy Soffer.
\newblock Numerical implementation of the multiscale and averaging methods for
  quasi periodic systems.
\newblock \emph{Computer Physics Communications}, 221:\penalty0 235--245, 2017.

\bibitem[Lea et~al.(2000)Lea, Allen, and Haine]{lea2000sensitivity}
Daniel~J Lea, Myles~R Allen, and Thomas~WN Haine.
\newblock Sensitivity analysis of the climate of a chaotic system.
\newblock \emph{Tellus A: Dynamic Meteorology and Oceanography}, 52\penalty0
  (5):\penalty0 523--532, 2000.

\bibitem[K{\"o}hl and Willebrand(2002)]{kohl2002adjoint}
Armin K{\"o}hl and J{\"u}rgen Willebrand.
\newblock An adjoint method for the assimilation of statistical characteristics
  into eddy-resolving ocean models.
\newblock \emph{Tellus A: Dynamic meteorology and oceanography}, 54\penalty0
  (4):\penalty0 406--425, 2002.

\bibitem[Yang and Schoenholz(2017)]{yang2017mean}
Greg Yang and Samuel~S Schoenholz.
\newblock Mean field residual networks: On the edge of chaos.
\newblock \emph{arXiv preprint arXiv:1712.08969}, 2017.

\bibitem[Hayou et~al.(2019)Hayou, Doucet, and Rousseau]{hayou2019impact}
Soufiane Hayou, Arnaud Doucet, and Judith Rousseau.
\newblock On the impact of the activation function on deep neural networks
  training.
\newblock In \emph{International conference on machine learning}, pages
  2672--2680. PMLR, 2019.

\bibitem[Parmas et~al.(2018)Parmas, Rasmussen, Peters, and
  Doya]{parmas2018pipps}
Paavo Parmas, Carl~Edward Rasmussen, Jan Peters, and Kenji Doya.
\newblock Pipps: Flexible model-based policy search robust to the curse of
  chaos.
\newblock In \emph{International Conference on Machine Learning}, pages
  4062--4071, 2018.

\bibitem[Parmas()]{Parmas_phdthesis}
Paavo Parmas.
\newblock Total stochastic gradient algorithms and applications to model-based
  reinforcement learning.

\bibitem[Metz et~al.(2019)Metz, Maheswaranathan, Nixon, Freeman, and
  Sohl-Dickstein]{metz2019understanding}
Luke Metz, Niru Maheswaranathan, Jeremy Nixon, Daniel Freeman, and Jascha
  Sohl-Dickstein.
\newblock Understanding and correcting pathologies in the training of learned
  optimizers.
\newblock In \emph{International Conference on Machine Learning}, pages
  4556--4565, 2019.

\bibitem[Ni and Wang(2017)]{ni2017sensitivity}
Angxiu Ni and Qiqi Wang.
\newblock Sensitivity analysis on chaotic dynamical systems by non-intrusive
  least squares shadowing (nilss).
\newblock \emph{Journal of Computational Physics}, 347:\penalty0 56--77, 2017.

\bibitem[Kochkov et~al.(2021)Kochkov, Smith, Alieva, Wang, Brenner, and
  Hoyer]{kochkov2021machine}
Dmitrii Kochkov, Jamie~A Smith, Ayya Alieva, Qing Wang, Michael~P Brenner, and
  Stephan Hoyer.
\newblock Machine learning--accelerated computational fluid dynamics.
\newblock \emph{Proceedings of the National Academy of Sciences}, 118\penalty0
  (21), 2021.

\bibitem[Ingraham et~al.(2018)Ingraham, Riesselman, Sander, and
  Marks]{ingraham2018learning}
John Ingraham, Adam Riesselman, Chris Sander, and Debora Marks.
\newblock Learning protein structure with a differentiable simulator.
\newblock In \emph{International Conference on Learning Representations}, 2018.

\bibitem[Ruelle(2009)]{ruelle2009review}
David Ruelle.
\newblock A review of linear response theory for general differentiable
  dynamical systems.
\newblock \emph{Nonlinearity}, 22\penalty0 (4):\penalty0 855, 2009.

\bibitem[Bollt(2000)]{bollt2000controlling}
Erik~M Bollt.
\newblock Controlling chaos and the inverse frobenius--perron problem: global
  stabilization of arbitrary invariant measures.
\newblock \emph{International Journal of Bifurcation and Chaos}, 10\penalty0
  (05):\penalty0 1033--1050, 2000.

\bibitem[Coluci et~al.(2005)Coluci, Legoas, De~Aguiar, and
  Galvao]{coluci2005chaotic}
VR~Coluci, SB~Legoas, MAM De~Aguiar, and DS~Galvao.
\newblock Chaotic signature in the motion of coupled carbon nanotube
  oscillators.
\newblock \emph{Nanotechnology}, 16\penalty0 (4):\penalty0 583, 2005.

\bibitem[Lorenz(1963)]{lorenz1963deterministic}
Edward~N Lorenz.
\newblock Deterministic nonperiodic flow.
\newblock \emph{Journal of atmospheric sciences}, 20\penalty0 (2):\penalty0
  130--141, 1963.

\bibitem[Kingma and Welling(2013)]{kingma2013auto}
Diederik~P Kingma and Max Welling.
\newblock Auto-encoding variational bayes.
\newblock \emph{arXiv preprint arXiv:1312.6114}, 2013.

\bibitem[Schulman et~al.(2015)Schulman, Heess, Weber, and
  Abbeel]{schulman2015gradient}
John Schulman, Nicolas Heess, Theophane Weber, and Pieter Abbeel.
\newblock Gradient estimation using stochastic computation graphs.
\newblock \emph{arXiv preprint arXiv:1506.05254}, 2015.

\bibitem[Ruelle et~al.(1980)]{ruelle1995measures}
David Ruelle et~al.
\newblock Measures describing a turbulent flow.
\newblock \emph{Annals of the New York Academy of Sciences}, page 357, 1980.

\bibitem[Gallavotti and Cohen(1995)]{gallavotti1995dynamical}
Giovanni Gallavotti and Ezechiel Godert~David Cohen.
\newblock Dynamical ensembles in stationary states.
\newblock \emph{Journal of Statistical Physics}, 80\penalty0 (5):\penalty0
  931--970, 1995.

\bibitem[Rechenberg(1973)]{rechenberg1973evolutionsstrategie}
Ingo Rechenberg.
\newblock Evolutionsstrategie--optimierung technisher systeme nach prinzipien
  der biologischen evolution.
\newblock 1973.

\bibitem[Schwefel(1977)]{schwefel1977evolutionsstrategien}
Hans-Paul Schwefel.
\newblock Evolutionsstrategien f{\"u}r die numerische optimierung.
\newblock In \emph{Numerische Optimierung von Computer-Modellen mittels der
  Evolutionsstrategie}, pages 123--176. Springer, 1977.

\bibitem[Wierstra et~al.(2008)Wierstra, Schaul, Peters, and
  Schmidhuber]{wierstra2008natural}
Daan Wierstra, Tom Schaul, Jan Peters, and Juergen Schmidhuber.
\newblock Natural evolution strategies.
\newblock In \emph{Evolutionary Computation, 2008. CEC 2008.(IEEE World
  Congress on Computational Intelligence). IEEE Congress on}, pages 3381--3387.
  IEEE, 2008.

\bibitem[Staines and Barber(2012)]{staines2012variational}
Joe Staines and David Barber.
\newblock Variational optimization.
\newblock \emph{arXiv preprint arXiv:1212.4507}, 2012.

\bibitem[Gal(2016)]{gal2016uncertainty}
Yarin Gal.
\newblock Uncertainty in deep learning.
\newblock 2016.

\bibitem[Mohamed et~al.(2020)Mohamed, Rosca, Figurnov, and
  Mnih]{mohamed2020monte}
Shakir Mohamed, Mihaela Rosca, Michael Figurnov, and Andriy Mnih.
\newblock Monte carlo gradient estimation in machine learning.
\newblock \emph{J. Mach. Learn. Res.}, 21\penalty0 (132):\penalty0 1--62, 2020.

\bibitem[Kong and Tao(2020)]{kong2020stochasticity}
Lingkai Kong and Molei Tao.
\newblock Stochasticity of deterministic gradient descent: Large learning rate
  for multiscale objective function.
\newblock \emph{arXiv preprint arXiv:2002.06189}, 2020.

\bibitem[Xiao et~al.(2020)Xiao, Pennington, and
  Schoenholz]{xiao2020disentangling}
Lechao Xiao, Jeffrey Pennington, and Samuel Schoenholz.
\newblock Disentangling trainability and generalization in deep neural
  networks.
\newblock In \emph{International Conference on Machine Learning}, pages
  10462--10472. PMLR, 2020.

\bibitem[Pearlmutter(1996)]{pearlmutter1996investigation}
Barak Pearlmutter.
\newblock \emph{An investigation of the gradient descent process in neural
  networks}.
\newblock PhD thesis, Carnegie Mellon University Pittsburgh, PA, 1996.

\bibitem[Maclaurin et~al.(2015)Maclaurin, Duvenaud, and
  Adams]{maclaurin2015gradient}
Dougal Maclaurin, David Duvenaud, and Ryan Adams.
\newblock Gradient-based hyperparameter optimization through reversible
  learning.
\newblock In \emph{International Conference on Machine Learning}, pages
  2113--2122, 2015.

\bibitem[LeCun(1998)]{lecun1998mnist}
Yann LeCun.
\newblock The mnist database of handwritten digits.
\newblock \emph{http://yann. lecun. com/exdb/mnist/}, 1998.

\bibitem[Lodi et~al.(2002)Lodi, Martello, and Monaci]{LODI2002241}
Andrea Lodi, Silvano Martello, and Michele Monaci.
\newblock Two-dimensional packing problems: A survey.
\newblock \emph{European Journal of Operational Research}, 141\penalty0
  (2):\penalty0 241--252, 2002.
\newblock ISSN 0377-2217.
\newblock \doi{https://doi.org/10.1016/S0377-2217(02)00123-6}.
\newblock URL
  \url{https://www.sciencedirect.com/science/article/pii/S0377221702001236}.

\bibitem[O'Hern et~al.(2003)O'Hern, Silbert, Liu, and Nagel]{ohern2003}
Corey~S. O'Hern, Leonardo~E. Silbert, Andrea~J. Liu, and Sidney~R. Nagel.
\newblock Jamming at zero temperature and zero applied stress: The epitome of
  disorder.
\newblock \emph{Phys. Rev. E}, 68:\penalty0 011306, Jul 2003.
\newblock \doi{10.1103/PhysRevE.68.011306}.
\newblock URL \url{https://link.aps.org/doi/10.1103/PhysRevE.68.011306}.

\bibitem[Bitzek et~al.(2006)Bitzek, Koskinen, G{\"a}hler, Moseler, and
  Gumbsch]{bitzek2006structural}
Erik Bitzek, Pekka Koskinen, Franz G{\"a}hler, Michael Moseler, and Peter
  Gumbsch.
\newblock Structural relaxation made simple.
\newblock \emph{Physical review letters}, 97\penalty0 (17):\penalty0 170201,
  2006.

\bibitem[Goodrich et~al.(2021)Goodrich, King, Schoenholz, Cubuk, and
  Brenner]{goodrich2021designing}
Carl~P Goodrich, Ella~M King, Samuel~S Schoenholz, Ekin~D Cubuk, and Michael~P
  Brenner.
\newblock Designing self-assembling kinetics with differentiable statistical
  physics models.
\newblock \emph{Proceedings of the National Academy of Sciences}, 118\penalty0
  (10), 2021.

\bibitem[Kaymak et~al.(2021)Kaymak, Rahnamoun, O'Hearn, van Duin, Merz~Jr, and
  Aktulga]{kaymak2021jax}
Mehmet~Cagri Kaymak, Ali Rahnamoun, Kurt~A O'Hearn, Adri~CT van Duin, Kenneth~M
  Merz~Jr, and Hasan~Metin Aktulga.
\newblock Jax-reaxff: A gradient based framework for extremely fast
  optimization of reactive force fields.
\newblock 2021.

\bibitem[Pascanu et~al.(2013)Pascanu, Mikolov, and
  Bengio]{pascanu2013difficulty}
Razvan Pascanu, Tomas Mikolov, and Yoshua Bengio.
\newblock On the difficulty of training recurrent neural networks.
\newblock In \emph{International Conference on Machine Learning}, pages
  1310--1318, 2013.

\bibitem[Le et~al.(2015)Le, Jaitly, and Hinton]{le2015simple}
Quoc~V Le, Navdeep Jaitly, and Geoffrey~E Hinton.
\newblock A simple way to initialize recurrent networks of rectified linear
  units.
\newblock \emph{arXiv preprint arXiv:1504.00941}, 2015.

\bibitem[Hochreiter and Schmidhuber(1997)]{hochreiter1997long}
Sepp Hochreiter and J{\"u}rgen Schmidhuber.
\newblock Long short-term memory.
\newblock \emph{Neural computation}, 9\penalty0 (8):\penalty0 1735--1780, 1997.

\bibitem[Chung et~al.(2014)Chung, Gulcehre, Cho, and
  Bengio]{chung2014empirical}
Junyoung Chung, Caglar Gulcehre, KyungHyun Cho, and Yoshua Bengio.
\newblock Empirical evaluation of gated recurrent neural networks on sequence
  modeling.
\newblock \emph{arXiv preprint arXiv:1412.3555}, 2014.

\bibitem[Collins et~al.(2016)Collins, Sohl-Dickstein, and
  Sussillo]{collins2016capacity}
Jasmine Collins, Jascha Sohl-Dickstein, and David Sussillo.
\newblock Capacity and trainability in recurrent neural networks.
\newblock \emph{arXiv preprint arXiv:1611.09913}, 2016.

\bibitem[Bayer(2015)]{bayer2015learning}
Justin~Simon Bayer.
\newblock \emph{Learning sequence representations}.
\newblock PhD thesis, Technische Universit{\"a}t M{\"u}nchen, 2015.

\bibitem[Monfared et~al.(2021)Monfared, Mikhaeil, and
  Durstewitz]{monfared2021train}
Zahra Monfared, Jonas~M Mikhaeil, and Daniel Durstewitz.
\newblock How to train rnns on chaotic data?
\newblock \emph{arXiv preprint arXiv:2110.07238}, 2021.

\bibitem[Huang et~al.(2021)Huang, Hu, Du, Zhou, Su, Tenenbaum, and
  Gan]{huang2021plasticinelab}
Zhiao Huang, Yuanming Hu, Tao Du, Siyuan Zhou, Hao Su, Joshua~B. Tenenbaum, and
  Chuang Gan.
\newblock Plasticinelab: A soft-body manipulation benchmark with differentiable
  physics.
\newblock 2021.

\bibitem[Cleac'h et~al.(2021)Cleac'h, Howell, Schwager, and
  Manchester]{cleach2021fast}
Simon~Le Cleac'h, Taylor Howell, Mac Schwager, and Zachary Manchester.
\newblock Fast contact-implicit model-predictive control.
\newblock 2021.

\bibitem[Ashcroft et~al.(1976)Ashcroft, Mermin, et~al.]{ashcroft1976solid}
Neil~W Ashcroft, N~David Mermin, et~al.
\newblock Solid state physics, 1976.

\bibitem[Truhlar et~al.(1996)Truhlar, Garrett, and
  Klippenstein]{truhlar1996current}
Donald~G Truhlar, Bruce~C Garrett, and Stephen~J Klippenstein.
\newblock Current status of transition-state theory.
\newblock \emph{The Journal of physical chemistry}, 100\penalty0 (31):\penalty0
  12771--12800, 1996.

\bibitem[Blondel et~al.(2021)Blondel, Berthet, Cuturi, Frostig, Hoyer,
  Llinares-López, Pedregosa, and Vert]{blondel2021efficient}
Mathieu Blondel, Quentin Berthet, Marco Cuturi, Roy Frostig, Stephan Hoyer,
  Felipe Llinares-López, Fabian Pedregosa, and Jean-Philippe Vert.
\newblock Efficient and modular implicit differentiation, 2021.

\bibitem[Rajeswaran et~al.(2019)Rajeswaran, Finn, Kakade, and
  Levine]{rajeswaran2019meta}
Aravind Rajeswaran, Chelsea Finn, Sham Kakade, and Sergey Levine.
\newblock Meta-learning with implicit gradients.
\newblock 2019.

\bibitem[Bai et~al.(2019)Bai, Kolter, and Koltun]{bai2019deep}
Shaojie Bai, J~Zico Kolter, and Vladlen Koltun.
\newblock Deep equilibrium models.
\newblock \emph{arXiv preprint arXiv:1909.01377}, 2019.

\bibitem[Abarbanel et~al.(2010)Abarbanel, Kostuk, and
  Whartenby]{abarbanel2010data}
Henry~DI Abarbanel, Mark Kostuk, and William Whartenby.
\newblock Data assimilation with regularized nonlinear instabilities.
\newblock \emph{Quarterly Journal of the Royal Meteorological Society: A
  journal of the atmospheric sciences, applied meteorology and physical
  oceanography}, 136\penalty0 (648):\penalty0 769--783, 2010.

\bibitem[Werbos(1990)]{werbos1990backpropagation}
Paul~J Werbos.
\newblock Backpropagation through time: what it does and how to do it.
\newblock \emph{Proceedings of the IEEE}, 78\penalty0 (10):\penalty0
  1550--1560, 1990.

\bibitem[Tallec and Ollivier(2017)]{tallec2017unbiasing}
Corentin Tallec and Yann Ollivier.
\newblock Unbiasing truncated backpropagation through time.
\newblock \emph{arXiv preprint arXiv:1705.08209}, 2017.

\bibitem[Sutskever(2013)]{sutskever2013training}
Ilya Sutskever.
\newblock \emph{Training recurrent neural networks}.
\newblock University of Toronto Toronto, Canada, 2013.

\bibitem[Andrychowicz et~al.(2016)Andrychowicz, Denil, Gomez, Hoffman, Pfau,
  Schaul, and de~Freitas]{andrychowicz2016learning}
Marcin Andrychowicz, Misha Denil, Sergio Gomez, Matthew~W Hoffman, David Pfau,
  Tom Schaul, and Nando de~Freitas.
\newblock Learning to learn by gradient descent by gradient descent.
\newblock In \emph{Advances in Neural Information Processing Systems}, pages
  3981--3989, 2016.

\bibitem[Wichrowska et~al.(2017)Wichrowska, Maheswaranathan, Hoffman,
  Colmenarejo, Denil, de~Freitas, and Sohl-Dickstein]{wichrowska2017learned}
Olga Wichrowska, Niru Maheswaranathan, Matthew~W Hoffman, Sergio~Gomez
  Colmenarejo, Misha Denil, Nando de~Freitas, and Jascha Sohl-Dickstein.
\newblock Learned optimizers that scale and generalize.
\newblock \emph{International Conference on Machine Learning}, 2017.

\bibitem[Lv et~al.(2017)Lv, Jiang, and Li]{lv2017learning}
Kaifeng Lv, Shunhua Jiang, and Jian Li.
\newblock Learning gradient descent: Better generalization and longer horizons.
\newblock \emph{arXiv preprint arXiv:1703.03633}, 2017.

\bibitem[Wu et~al.(2016)Wu, Ren, Liao, and Grosse]{wuunderstanding}
Yuhuai Wu, Mengye Ren, Renjie Liao, and Roger~B Grosse.
\newblock Understanding short-horizon bias in stochastic meta-optimization.
\newblock pages 478--487, 2016.

\bibitem[Kaiser and Sutskever(2015)]{kaiser2015neural}
{\L}ukasz Kaiser and Ilya Sutskever.
\newblock Neural gpus learn algorithms.
\newblock \emph{arXiv preprint arXiv:1511.08228}, 2015.

\bibitem[Merity et~al.(2017)Merity, Keskar, and Socher]{merity2017regularizing}
Stephen Merity, Nitish~Shirish Keskar, and Richard Socher.
\newblock Regularizing and optimizing lstm language models.
\newblock \emph{arXiv preprint arXiv:1708.02182}, 2017.

\bibitem[Gehring et~al.(2017)Gehring, Auli, Grangier, Yarats, and
  Dauphin]{gehring2017convolutional}
Jonas Gehring, Michael Auli, David Grangier, Denis Yarats, and Yann~N Dauphin.
\newblock Convolutional sequence to sequence learning.
\newblock In \emph{International Conference on Machine Learning}, pages
  1243--1252. PMLR, 2017.

\bibitem[Pilyugin(2006)]{pilyugin2006shadowing}
Sergei~Yu Pilyugin.
\newblock \emph{Shadowing in dynamical systems}.
\newblock Springer, 2006.

\bibitem[Wang et~al.(2014)Wang, Hu, and Blonigan]{wang2014least}
Qiqi Wang, Rui Hu, and Patrick Blonigan.
\newblock Least squares shadowing sensitivity analysis of chaotic limit cycle
  oscillations.
\newblock \emph{Journal of Computational Physics}, 267:\penalty0 210--224,
  2014.

\bibitem[Ni et~al.(2019)Ni, Wang, Fernandez, and Talnikar]{ni2019sensitivity}
Angxiu Ni, Qiqi Wang, Pablo Fernandez, and Chaitanya Talnikar.
\newblock Sensitivity analysis on chaotic dynamical systems by finite
  difference non-intrusive least squares shadowing (fd-nilss).
\newblock \emph{Journal of Computational Physics}, 394:\penalty0 615--631,
  2019.

\bibitem[Ni(2021)]{ni2021numerical}
Angxiu Ni.
\newblock \emph{Numerical Differentiation of Stationary Measures of Chaos}.
\newblock PhD thesis, University of California, Berkeley, 2021.

\bibitem[Blonigan et~al.(2014)Blonigan, Gomez, and Wang]{blonigan2014least}
Patrick~J Blonigan, Steven~A Gomez, and Qiqi Wang.
\newblock Least squares shadowing for sensitivity analysis of turbulent fluid
  flows.
\newblock In \emph{52nd Aerospace Sciences Meeting}, page 1426, 2014.

\bibitem[Blonigan et~al.(2017)Blonigan, Fernandez, Murman, Wang, Rigas, and
  Magri]{blonigan2017toward}
Patrick~J Blonigan, Pablo Fernandez, Scott~M Murman, Qiqi Wang, Georgios Rigas,
  and Luca Magri.
\newblock Toward a chaotic adjoint for les.
\newblock \emph{arXiv preprint arXiv:1702.06809}, 2017.

\bibitem[Wang(2013)]{wang2013forward}
Qiqi Wang.
\newblock Forward and adjoint sensitivity computation of chaotic dynamical
  systems.
\newblock \emph{Journal of Computational Physics}, 235:\penalty0 1--13, 2013.

\bibitem[Thuburn(2005)]{thuburn2005climate}
J~Thuburn.
\newblock Climate sensitivities via a fokker--planck adjoint approach.
\newblock \emph{Quarterly Journal of the Royal Meteorological Society: A
  journal of the atmospheric sciences, applied meteorology and physical
  oceanography}, 131\penalty0 (605):\penalty0 73--92, 2005.

\bibitem[Abramov and Majda(2007)]{abramov2007blended}
Rafail~V Abramov and Andrew~J Majda.
\newblock Blended response algorithms for linear fluctuation-dissipation for
  complex nonlinear dynamical systems.
\newblock \emph{Nonlinearity}, 20\penalty0 (12):\penalty0 2793, 2007.

\bibitem[Cooper and Haynes(2011)]{cooper2011climate}
Fenwick~C Cooper and Peter~H Haynes.
\newblock Climate sensitivity via a nonparametric fluctuation--dissipation
  theorem.
\newblock \emph{Journal of the Atmospheric Sciences}, 68\penalty0 (5):\penalty0
  937--953, 2011.

\bibitem[Guti{\'e}rrez and Lucarini(2019)]{gutierrez2019response}
Manuel~Santos Guti{\'e}rrez and Valerio Lucarini.
\newblock Response and sensitivity using markov chains.
\newblock \emph{arXiv preprint arXiv:1907.12881}, 2019.

\bibitem[Chandramoorthy and Wang(2020)]{chandramoorthy2020computable}
Nisha Chandramoorthy and Qiqi Wang.
\newblock A computable realization of ruelle's formula for linear response of
  statistics in chaotic systems.
\newblock \emph{arXiv preprint arXiv:2002.04117}, 2020.

\bibitem[Chandramoorthy and Wang(2021)]{chandramoorthy2021efficient}
Nisha Chandramoorthy and Qiqi Wang.
\newblock Efficient computation of linear response of chaotic attractors with
  one-dimensional unstable manifolds.
\newblock \emph{arXiv preprint arXiv:2103.08816}, 2021.

\bibitem[Chandramoorthy(2021)]{chandramoorthy21thesis}
Nisha Chandramoorthy.
\newblock \emph{An efficient algorithm for sensitivity analysis of chaotic
  systems}.
\newblock PhD thesis, Massachusetts Institute of Technology, 2021.
\newblock URL
  \url{https://web.mit.edu/nishac/www/papers/PhD_Thesis-compressed.pdf}.

\bibitem[Ladick{\`y} et~al.(2015)Ladick{\`y}, Jeong, Solenthaler, Pollefeys,
  and Gross]{ladicky2015data}
L'ubor Ladick{\`y}, SoHyeon Jeong, Barbara Solenthaler, Marc Pollefeys, and
  Markus Gross.
\newblock Data-driven fluid simulations using regression forests.
\newblock \emph{ACM Transactions on Graphics (TOG)}, 34\penalty0 (6):\penalty0
  1--9, 2015.

\bibitem[Tompson et~al.(2017)Tompson, Schlachter, Sprechmann, and
  Perlin]{tompson2017accelerating}
Jonathan Tompson, Kristofer Schlachter, Pablo Sprechmann, and Ken Perlin.
\newblock Accelerating eulerian fluid simulation with convolutional networks.
\newblock In \emph{International Conference on Machine Learning}, pages
  3424--3433. PMLR, 2017.

\bibitem[Ha and Schmidhuber(2018)]{ha2018world}
David Ha and J{\"u}rgen Schmidhuber.
\newblock World models.
\newblock \emph{arXiv preprint arXiv:1803.10122}, 2018.

\bibitem[Hafner et~al.(2019)Hafner, Lillicrap, Fischer, Villegas, Ha, Lee, and
  Davidson]{hafner2019learning}
Danijar Hafner, Timothy Lillicrap, Ian Fischer, Ruben Villegas, David Ha,
  Honglak Lee, and James Davidson.
\newblock Learning latent dynamics for planning from pixels.
\newblock In \emph{International Conference on Machine Learning}, pages
  2555--2565. PMLR, 2019.

\bibitem[Ozair et~al.(2021)Ozair, Li, Razavi, Antonoglou, Oord, and
  Vinyals]{ozair2021vector}
Sherjil Ozair, Yazhe Li, Ali Razavi, Ioannis Antonoglou, A{\"a}ron van~den
  Oord, and Oriol Vinyals.
\newblock Vector quantized models for planning.
\newblock \emph{arXiv preprint arXiv:2106.04615}, 2021.

\bibitem[Feinberg et~al.(2018)Feinberg, Wan, Stoica, Jordan, Gonzalez, and
  Levine]{feinberg2018model}
Vladimir Feinberg, Alvin Wan, Ion Stoica, Michael~I Jordan, Joseph~E Gonzalez,
  and Sergey Levine.
\newblock Model-based value expansion for efficient model-free reinforcement
  learning.
\newblock In \emph{Proceedings of the 35th International Conference on Machine
  Learning (ICML 2018)}, 2018.

\bibitem[Buckman et~al.(2018)Buckman, Hafner, Tucker, Brevdo, and
  Lee]{buckman2018sample}
Jacob Buckman, Danijar Hafner, George Tucker, Eugene Brevdo, and Honglak Lee.
\newblock Sample-efficient reinforcement learning with stochastic ensemble
  value expansion.
\newblock \emph{arXiv preprint arXiv:1807.01675}, 2018.

\bibitem[Kurutach et~al.(2018)Kurutach, Clavera, Duan, Tamar, and
  Abbeel]{kurutach2018model}
Thanard Kurutach, Ignasi Clavera, Yan Duan, Aviv Tamar, and Pieter Abbeel.
\newblock Model-ensemble trust-region policy optimization.
\newblock \emph{arXiv preprint arXiv:1802.10592}, 2018.

\bibitem[Clavera et~al.(2020)Clavera, Fu, and Abbeel]{clavera2020model}
Ignasi Clavera, Violet Fu, and Pieter Abbeel.
\newblock Model-augmented actor-critic: Backpropagating through paths.
\newblock \emph{arXiv preprint arXiv:2005.08068}, 2020.

\bibitem[Williams(1992)]{williams1992simple}
Ronald~J Williams.
\newblock Simple statistical gradient-following algorithms for connectionist
  reinforcement learning.
\newblock \emph{Machine learning}, 8\penalty0 (3-4):\penalty0 229--256, 1992.

\bibitem[Salimans et~al.(2017)Salimans, Ho, Chen, Sidor, and
  Sutskever]{salimans2017evolution}
Tim Salimans, Jonathan Ho, Xi~Chen, Szymon Sidor, and Ilya Sutskever.
\newblock Evolution strategies as a scalable alternative to reinforcement
  learning.
\newblock \emph{arXiv preprint arXiv:1703.03864}, 2017.

\bibitem[Schulman et~al.(2017)Schulman, Wolski, Dhariwal, Radford, and
  Klimov]{schulman2017proximal}
John Schulman, Filip Wolski, Prafulla Dhariwal, Alec Radford, and Oleg Klimov.
\newblock Proximal policy optimization algorithms.
\newblock \emph{arXiv preprint arXiv:1707.06347}, 2017.

\bibitem[Griewank and Walther(2000)]{griewank2000algorithm}
Andreas Griewank and Andrea Walther.
\newblock Algorithm 799: revolve: an implementation of checkpointing for the
  reverse or adjoint mode of computational differentiation.
\newblock \emph{ACM Transactions on Mathematical Software (TOMS)}, 26\penalty0
  (1):\penalty0 19--45, 2000.

\bibitem[Chen et~al.(2016)Chen, Xu, Zhang, and Guestrin]{chen2016training}
Tianqi Chen, Bing Xu, Chiyuan Zhang, and Carlos Guestrin.
\newblock Training deep nets with sublinear memory cost.
\newblock \emph{arXiv preprint arXiv:1604.06174}, 2016.

\bibitem[Kawaguchi(2016)]{kawaguchi2016deep}
Kenji Kawaguchi.
\newblock Deep learning without poor local minima.
\newblock \emph{arXiv preprint arXiv:1605.07110}, 2016.

\bibitem[Kargin(2010)]{kargin2010products}
Vladislav Kargin.
\newblock Products of random matrices: Dimension and growth in norm.
\newblock \emph{The Annals of Applied Probability}, 20\penalty0 (3):\penalty0
  890--906, 2010.

\bibitem[Bajovic(2013)]{bajovic2013large}
Dragana Bajovic.
\newblock \emph{Large Deviations Rates for Distributed Inference}.
\newblock PhD thesis, Instituto Superior T{\'e}cnico Lisbon, Portugal, 2013.

\end{thebibliography}
\bibliographystyle{unsrtnat}

\appendix
\clearpage

\section{Progression of dynamical systems} \label{Appendix}

Dynamical progression of training systems are very subtle. In it's
core is an iterative algebraic process where one defines a canonical
transformation to take a state vector from time $t$ to $t+1$. Albiet
for long range temporal processes there could be a cascade of confounder
that effect the convergence properties or the generated dynamics.
In the context of automatic differentiation this can become even more
critical since the typical number of dynamical steps spans vast orders
of magnitude. In what follow we will overview some of the convergence,
or lack of, properties for different scenarios of the dynamical progression.

\subsection{Deterministic single transformation}

We start off by considering the iterative map $A$ as a propagator
of dynamics i.e 
\[
x^{1}=Ax^{0}
\]
where the supscript denotes at time step. If the transformation is
always the same then the state at point $k$ is simply 
\begin{equation}
x^{k}=A^{k}x_{0}\label{eq:him_dynamica_eval}
\end{equation}
we note, that this will only converge for every initial state $x_{0}$
iif the eigenvalues $\lambda\ne1$ have an amplitude $\left|\lambda\right|<1$
and more harshly if the spectrum contains $\lambda^{*}=1$ then it
has to have a rank that fulfills $\phi\left(\lambda\right)=\det\left(\lambda I-A\right)$.
To show this, let us examine thee Jordan canonical form of the matrix
\[
A=XDX^{-1}
\]
 where $X$ is a matrix who's columns are the eigenvalues and $D$
is a diagonal matrix who's diagonal are the eigenvalues. The dynamical
progression \ref{eq:him_dynamica_eval} under the Jordan form becomes
\[
x^{k}=A^{k}x_{0}=\left(XDX^{-1}\right)^{k}x_{0}=X\left(\begin{array}{cccc}
\lambda_{1}^{k}\\
 & \lambda_{2}^{k}\\
 &  & \vdots\\
 &  &  & \lambda_{n}^{k}
\end{array}\right)X^{-1}x_{0}
\]
 we see that any eigenvalues smaller then 1 will decay with larger
maps, and eigenvalues which are bigger then one will diverge. Interestingly,
if the eigenvalue of $\lambda=1$ does exist, such eigenvalues may
yield periodic or nearly periodic orbits or even orbits that may diverge
to infinity. 

\subsection{Deterministic multiple transformation}

\begin{figure}
\begin{centering}
\includegraphics[scale=0.4]{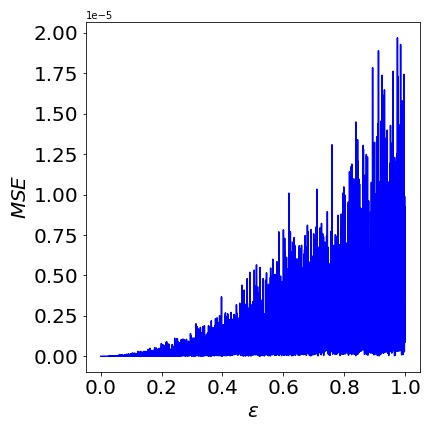}
\par\end{centering}
\caption{\label{fig:MSE-as-a}MSE as a function of a small perturbation $\epsilon\in\mathbb{R}^{N\times N}$
where $X^{i+1}=X^{i}+\epsilon$ }

\end{figure}

Last section all of the time dynamics was exactly the same, we can
relax this assumption by looking at transformation functions that
are inhomoginues over time, i.e.,
\[
A^{i}\neq A^{i+1}
\]
 and thus the spectrum also changes over time. For this scenario we
can write down the dynamics in the following way 
\[
x^{k}=\prod_{i=0}^{k}A^{i}x_{0}
\]
and using the Jordan form 
\[
x^{k}=X^{0}\prod_{i=0}^{k}D^{i}\left(X^{i}\right)^{-1}x_{0}
\]
to simplify this lets look at first consider the case of small increments
between two steps $X^{i}D^{i}\left(X^{i}\right)^{-1}X^{i+1}D^{i+1}\left(X^{i+1}\right)^{-1}$.
For a slowly varying dynamics or matrices then we can approximate
\[
\left(X^{i}\right)^{-1}X^{i+1}\approx I+\epsilon
\]
Where we can numerically justify the small variation angle by a simple
numerical check as shown in figure \ref{fig:MSE-as-a}. With this
the transformation now becomes 

\[
x^{k}=X^{0}\prod_{i=0}^{k}D^{i}\left(X^{i}\right)^{-1}x_{0}=X^{0}\prod_{i=0}^{k}D^{0}\left(I+\epsilon\right)D^{1}\left(I+\epsilon\right)D^{2}......\left(X^{0}\right)^{-1}x_{0}
\]
\[
x^{k}\approx X^{0}\left(\prod_{i=0}^{k}D^{I}\right)\left(X^{0}\right)^{-1}x_{0}
\]
 
\[
x^{k}=X\left(\begin{array}{cccc}
\prod_{i=0}^{k}\lambda_{1}^{k}\\
 & \prod_{i=0}^{k}\lambda_{2}^{k}\\
 &  & \vdots\\
 &  &  & \prod_{i=0}^{k}\lambda_{n}^{k}
\end{array}\right)X^{-1}x_{0}
\]
such we see the same result, for eigenvalues which are bigger then
$1$ we will have divergence, for smaller then one the dynamic will
halt. More interestingly for this case, we don't just need a single
eigenvalue of $1$ but rather a cascade of same scale eigenvalues
for the dynamics to converge.

\subsection{Stochastic transformations }

More interestingly one can also consider the case of a random transformation
where $A$ is in fact a stochastic random matrix. For this case, let
us look at two different cases.

\subsubsection{Gaussian randomness}

For this case, it is easier to look at the product of eigenvalues
i.e the determinant of 
\[
x^{k}=\prod_{i=0}^{k}A^{i}x_{0}
\]
 and defining 
\[
d_{k}=\det\left(\prod_{i=0}^{k}A^{i}\right)=\prod_{i=0}^{k}\det A^{i}
\]
now if $a_{ij}\left(i\right)\sim\mathcal{N}\left(0,\sigma^{2}\right)$
then the expectation is 
\[
\mathbb{E}\left[d_{k}\right]=0
\]
which we also mark as
\[
\mathbb{E}\left[d_{k}\right]=\mu_{k}
\]
 now be denoting the symmetric and non symmetric parts of the minor
as 
\[
\prod_{i,j}a_{ij}\left(i\right)=\begin{cases}
S_{1}\left(i\right) & i=j\\
S_{2}\left(i\right) & i\neq j
\end{cases}
\]
Where $S$ are the chi-squared distribution, then we can write the
Variance as 
\[
\mathbb{V}\left[d_{k}\right]=\mathbb{E}\left[d_{k}^{2}\right]=\text{\ensuremath{\sigma^{4k}}\ensuremath{\prod_{i=0}^{k}}\ensuremath{\mathbb{E}}\ensuremath{\left[\left(S_{1}\left(i\right)-S_{2}\left(i\right)\right)^{2}\right]}}
\]
This is important since the difference of two chi squared distribution
is a generalized Laplace distribution
\[
S_{1}\left(i\right)-S_{2}\left(i\right)\sim\Gamma_{\nu}\left(\mu=0,\alpha=\frac{1}{2},\beta=0,\lambda=\frac{1}{2},\gamma=\frac{1}{2}\right)
\]
 which also tells us that 
\[
\mathbb{V}\left[S_{1}\left(i\right)-S_{2}\left(i\right)\right]=\frac{2\lambda}{\gamma^{2}}\left(1+\frac{2\beta^{2}}{\gamma^{2}}\right)=4
\]
 such 
\[
\mathbb{V}\left[d_{k}\right]=4^{k}\ensuremath{\sigma^{4k}}
\]
Similar to what we have seen before, the convergence of this depends
on the variance.

The problem of bounding non Gaussian noise has been extensively discussed in the literature \citep{kargin2010products,bajovic2013large}. While it is an interesting case, an in depth evaluation of this type bound is out of the scope of this paper and is the basis for future exploration. 

\subsubsection{On the special case of discrete laplacian }

We have discussed gradients and maps with their dynamical progression.
It's worthwhile to look at one special case, where the dynamical progression
is dependent on the hessian i.e a mapping of the following form 
\[
u_{t}=\alpha u_{xx}
\]
under some boundary conditions 
\[
\begin{cases}
x\in\left[a,b\right]\\
u\left(x,0\right)=f\left(x\right)\\
u\left(a,t\right)=u\left(b,t\right)=0
\end{cases}
\]
under a fixed grid discretization of $\Delta x=\frac{b-a}{n}$ the
dynamical process actually becomes 
\[
\left(\begin{array}{c}
u_{1}^{j+1}\\
u_{2}^{j+1}\\
\vdots\\
\vdots\\
u_{n-1}^{j+1}
\end{array}\right)=\left(\begin{array}{ccccc}
1-2\frac{\alpha\Delta t}{\Delta x^{2}} & \frac{\alpha\Delta t}{\Delta x^{2}}\\
\frac{\alpha\Delta t}{\Delta x^{2}} & 1-2\frac{\alpha\Delta t}{\Delta x^{2}}\\
 &  & \ddots\\
 &  &  & \ddots & \frac{\alpha\Delta t}{\Delta x^{2}}\\
 &  &  & \frac{\alpha\Delta t}{\Delta x^{2}} & 1-2\frac{\alpha\Delta t}{\Delta x^{2}}
\end{array}\right)\left(\begin{array}{c}
u_{1}^{j}\\
u_{2}^{j}\\
\vdots\\
\vdots\\
u_{n-1}^{j}
\end{array}\right)
\]
\[
u^{j}=A^{j}u^{0}
\]
If $\frac{\alpha\Delta t}{\Delta x^{2}}<0.5$ then the map matrix
$A$ is non negative and thus the forbeniues perron theorem tells
us that there is a dominant eigenvalue $\lambda>0$ with a non negative
eigenvector $\text{v}$. Building on this we can write the eigenvalue
equation (in coordinate to coordinate transformation) 
\[
\lambda v=Av
\]
\[
\lambda v_{i}=\frac{\alpha\Delta t}{\Delta x^{2}}v_{i-1}+\left(1-2\frac{\alpha\Delta t}{\Delta x^{2}}\right)v_{i}+hv_{i+1}\leq1
\]
 more so, if $i$ index corresponds to $v_{i}=1$ then $\lambda<1$.
This actually lets us state Gerschgorin\textquoteright s theorem:

Each eigenvalue $\lambda$ of the real or complex matrix $A=\left(a_{ij}\right)$
lies in at least one of the close disks 
\[
\left|\lambda-a_{ii}\right|\leq\sum_{j\neq i}\left|a_{ij}\right|
\]
 
 \subsubsection{On the special case of discrete drift diffusion }

As a further escalation, let us now look at the drift diffusion equation
\[
u_{t}=\alpha u_{xx}+\beta u_{x}
\]
under some boundary conditions 
\[
\begin{cases}
x\in\left[a,b\right]\\
u\left(x,0\right)=f\left(x\right)\\
u\left(a,t\right)=u\left(b,t\right)=0
\end{cases}
\]
under a fixed grid discretization of $\Delta x=\frac{b-a}{n}$ the
dynamical process actually becomes
\[
u_{i}^{j+1}=u_{i}^{j}+\frac{\alpha\Delta t}{\Delta x^{2}}\left(u_{i+1}^{j}-2u_{i}^{j}+u_{i-1}^{j}\right)+\frac{\beta\Delta t}{\Delta x}\left(\text{\ensuremath{u_{i+1}^{j}}-\ensuremath{u_{i}^{j}}}\right)
\]
\begin{align*}
\left(\begin{array}{c}
u_{1}^{j+1}\\
u_{2}^{j+1}\\
\vdots\\
\vdots\\
u_{n-1}^{j+1}
\end{array}\right) & =\left[\left(\begin{array}{ccccc}
1-2\frac{\alpha\Delta t}{\Delta x^{2}} & \frac{\alpha\Delta t}{\Delta x^{2}}\\
\frac{\alpha\Delta t}{\Delta x^{2}} & 1-2\frac{\alpha\Delta t}{\Delta x^{2}} & \frac{\alpha\Delta t}{\Delta x^{2}}\\
 &  & \ddots\\
 &  &  & \ddots & \frac{\alpha\Delta t}{\Delta x^{2}}\\
 &  &  & \frac{\alpha\Delta t}{\Delta x^{2}} & 1-2\frac{\alpha\Delta t}{\Delta x^{2}}
\end{array}\right)+\right.\\
 & \left.\left(\begin{array}{ccccc}
-\frac{\beta\Delta t}{\Delta x} & \frac{\beta\Delta t}{\Delta x}\\
0 & -\frac{\beta\Delta t}{\Delta x} & \frac{\beta\Delta t}{\Delta x}\\
 &  & \ddots\\
 &  &  & \ddots & \frac{\alpha\Delta t}{\Delta x^{2}}\\
 &  &  &  & \frac{\beta\Delta t}{\Delta x}
\end{array}\right)\right]\left(\begin{array}{c}
u_{1}^{j}\\
u_{2}^{j}\\
\vdots\\
\vdots\\
u_{n-1}^{j}
\end{array}\right)
\end{align*}
 In this case for the map to converge we have to require 
\[
1-2\frac{\alpha\Delta t}{\Delta x^{2}}-\frac{\beta\Delta t}{\Delta x}\geq0
\]
 
\[
\Delta x^{2}\frac{1}{\Delta t}-\Delta x\frac{\beta}{\Delta t}-\frac{2\alpha}{\Delta t}\geq0
\]
\[
\left(\Delta x-\frac{\beta}{\Delta t}\pm\sqrt{\beta^{2}+8\alpha}\right)\left(\Delta x+\frac{\beta}{\Delta t}\pm\sqrt{\beta^{2}+8\alpha}\right)\geq0
\]

\end{document}